\documentclass{article}
\usepackage[main,final]{neurips_2026}

\usepackage[utf8]{inputenc} 
\usepackage[T1]{fontenc}    
\usepackage{hyperref}       
\usepackage{url}            
\usepackage{booktabs}       
\usepackage{amsfonts}       
\usepackage{nicefrac}       
\usepackage{microtype}      
\usepackage{xcolor}         
\usepackage{graphicx}

\RequirePackage{tikz}
\usetikzlibrary{arrows.meta, shapes.geometric, positioning, 
                matrix, decorations.pathreplacing, 
                patterns, calc, backgrounds, fit, 
                shadows, intersections}

\RequirePackage{graphicx}
\RequirePackage{subcaption}
\RequirePackage{caption}
\RequirePackage{float}
\RequirePackage{placeins}

\RequirePackage{fontawesome5}

\newcommand{\ind}[1]{\mathbf{1}\!\left[#1\right]}
\RequirePackage{xcolor}
\RequirePackage{colortbl}
\RequirePackage{multirow}
\RequirePackage{booktabs}

\RequirePackage{amsmath, amssymb}
\RequirePackage{bm}

\usepackage{amsmath}      
\usepackage{amssymb}      
\usepackage{amsfonts}     
\usepackage{amsthm}       

\newtheorem{theorem}{Theorem}[section]          
\newtheorem{lemma}[theorem]{Lemma}              
\newtheorem{corollary}[theorem]{Corollary}      
\newtheorem{proposition}[theorem]{Proposition}  
\newtheorem{assumption}[theorem]{Assumption}

\usepackage{algorithm}
\usepackage{algorithmic}

\usepackage[most]{tcolorbox}
\tcbuselibrary{listings,breakable,skins}
\tcbset{
  promptbox/.style={
    enhanced,
    breakable,
    colback=gray!3,
    colframe=black!35,
    boxrule=0.6pt,
    arc=2pt,
    left=6pt,right=6pt,top=6pt,bottom=6pt,
    listing only,
    listing options={
      basicstyle=\ttfamily\footnotesize,
      breaklines=true,
      breakatwhitespace=false,
      columns=fullflexible,
      keepspaces=true,
      showstringspaces=false
    }
  }
}

\usepackage{subcaption} 

\usepackage{wrapfig}
\usepackage{xcolor}
\usepackage{colortbl}

\usepackage{fontawesome5}

\definecolor{HGUrow}{HTML}{F0E8FA}
\newcommand{\best}[1]{\textbf{#1}}
\newcommand{\second}[1]{\underline{#1}}

\newcommand{\drop}[1]{\textcolor{blue}{\scriptsize\,(\,--#1\,)}}

\title{Hallucination-Guided Unlearning: Using Hallucination Traces to Reveal Overfitted Memories}

\author{%
  Jiayu Zhang$^{1}$\thanks{Equal contribution.} \quad Ziqi Zhong$^{2}$\footnotemark[1] \quad
  Yuliang Gai$^{3,4}$ \quad Canran Xiao$^{5}$\thanks{Corresponding author: \texttt{xiaocr3@mail.sysu.edu.cn}.} \\
  $^{1}$Peking University \qquad $^{2}$UTS Business School, University of Technology Sydney \\
  $^{3}$Massachusetts Institute of Technology \qquad $^{4}$University of Massachusetts Boston \\
  $^{5}$Shenzhen Campus of Sun Yat-sen University
}
\hypersetup{
  hidelinks,
  pdftitle={Hallucination-Guided Unlearning: Using Hallucination Traces to Reveal Overfitted Memories},
  pdfauthor={Jiayu Zhang, Ziqi Zhong, Yuliang Gai, Canran Xiao},
  pdfsubject={NeurIPS 2026 camera-ready paper},
  pdfkeywords={large language models, machine unlearning, hallucination tracing}
}

\begin{document}

\maketitle

\begin{abstract}
Selective unlearning is increasingly important for the safety and compliance of large language models. Harmful or memorized content is often entangled with useful knowledge and difficult to enumerate. We use the model’s own failures to identify candidates for unlearning.
We propose \textsc{Hallucination-Guided Unlearning} (HGU), which uses hallucinations as diagnostic signals. HGU localizes unsupported spans via evidence and internal consensus, traces a sparse set of high-responsibility MLP neurons, and edits them with compact gates under a utility-preserving KL constraint.
On TOFU, HGU achieves the best MU--FE trade-off, with Avg scores of 0.7221 (forget05) and 0.7699 (forget10), and the highest SepS H-Avg (0.4382/0.4715) under mixed forget/retain prompts. On the WMDP auxiliary task, HGU yields the strongest forgetting (e.g., 14.2 on Economics) while improving Retain (52.7) and MMLU (57.2).
These results suggest that tracing hallucinations enables targeted, post-hoc unlearning that better balances forgetting efficacy and general utility without full retraining.
\end{abstract}

\section{Introduction}
\label{sec:intro}

Large language models (LLMs) are increasingly deployed as knowledge-intensive interfaces for search, coding, education, and decision support, where users expect both high utility and responsible behavior \citep{xiao2026cora,xiao2026yang}.
Medical decision-making and financial trading, for example, have motivated safeguards that check the evidence used by LLM agents \citep{hu2026research,wang2026tradingagentsmarketcrashers}.
Yet training pipelines inevitably imprint undesirable content into model parameters---ranging from private or copyrighted text to hazardous know-how---making selective removal a first-class requirement for safety, compliance, and trustworthy deployment \citep{Bourtoule2021,Li2024}.
Because memorized influence can be diffuse and entangled with legitimate knowledge, unlearning must balance effective forgetting against broad capability preservation under tight compute constraints \citep{Maini2024,Shi2024MUSE,yang2026privacyalignment}.

Recent work has refined unlearning objectives and interventions and developed benchmarks for measuring forgetting--utility trade-offs \citep{Maini2024,Li2024,Shi2024MUSE}.
Most approaches optimize removal given an \emph{explicit target}, such as a curated forget set, task vector, or labeled harmful corpus \citep{ZhangR2024,Ji2024,Gao2024,Hu2025,Shen2025,ZhangS2025,Zhao2025}.
When memorized content is \emph{latent}, triggered only by certain prompts, or mixed with benign facts, unlearning also requires \emph{discovering what to forget} and isolating the responsible internal pathways \citep{Shi2024MUSE,Li2024}.
Confident hallucinations can reveal overfitted memories that lack evidential support. We use these generation failures to guide unlearning.

\definecolor{AppBlue}{HTML}{7C3AED}    
\definecolor{AppBlueBg}{HTML}{F3E8FF}  

\begin{tcolorbox}[
  enhanced,
  colback=AppBlueBg,
  colframe=AppBlue,
  colbacktitle=AppBlue,   
  coltitle=white,         
  fonttitle=\bfseries,
  title=\faIcon{question} This paper asks:,
  boxrule=0.8pt,
  arc=2.5mm,
  left=10pt,right=10pt,top=8pt,bottom=8pt,
]
\emph{  Can hallucinations help us find what the model has memorized incorrectly, and remove it without retraining?}
\end{tcolorbox}
We compare generated content with trusted evidence and measure consensus across samples to localize hallucinated spans. We trace those spans to a sparse set of memory pathways and edit the pathways under a utility-preserving constraint.

\begin{itemize}
  \item[\textbf{(i)}] We formulate unsupported-span-guided unlearning, using hallucinated spans as diagnostic probes.
  \item[\textbf{(ii)}] We propose HGU, a sparse post-hoc editing method that combines evidence-based span localization, exact local activation-gradient tracing, and utility-preserving gated intervention.
  \item[\textbf{(iii)}] We report leakage-free protocols and attribution implementation details. We evaluate on TOFU (including the SepS mixed-prompt protocol), MUSE, WMDP, and PKU-SafeRLHF, and report seed variability, compute costs, and quantization robustness.
\end{itemize}

\textbf{Scope.}
HGU uses unsupported spans to identify candidates for reference- or policy-conditioned unlearning. Selecting accurate content for removal requires a policy or provenance signal. Missing knowledge and retrieval errors may call for correction.
The evidence needed depends on the task, as when pointing gestures help disambiguate language in embodied reference resolution \citep{Li2023TouchLine}.
The selected channels serve as intervention points, with the same knowledge potentially represented elsewhere.
Appendix~\ref{app:review_evidence} reports blind discovery, independent evaluation, and re-elicitation tests for approximate post-hoc unlearning.

\section{Related Work}
\label{sec:related_work}

\textbf{LLM unlearning and evaluation.}
In federated learning, training-time influence logging facilitates client-level deletion \citep{Xiao2026IDFT}.
LLM unlearning is commonly posed as removing the influence of a specified \emph{forget set} while preserving utility on \emph{retain} data and downstream tasks, extending classical machine unlearning to generative models \citep{Bourtoule2021}.
Recent benchmarks clarify this objective and reveal gaps in practice: \textsc{TOFU} tests fictitious profile forgetting \citep{Maini2024}, \textsc{MUSE} evaluates real-corpus removal across memorization, privacy, and deployment constraints \citep{Shi2024MUSE}, and \textsc{WMDP} frames unlearning as safety-motivated suppression of hazardous knowledge \citep{Li2024}.
Algorithmically, prior work refines ``forget-vs.-retain'' objectives to avoid collapse (e.g., NPO and related gradient-based or multi-objective variants) \citep{ZhangR2024,Maini2024,Pan2025}, while logit-space approaches such as ULD edit behavior via learned logit differences \citep{Ji2024}.
Most methods assume explicit targets. Their performance degrades when harmful influence is latent, entangled, or triggered only by particular prompts, a limitation highlighted by broad, sequential evaluations \citep{Shi2024MUSE}.
We address this upstream bottleneck by using hallucinations to \emph{identify} what to forget and performing sparse, trace-guided edits under an explicit utility-drift constraint.

\textbf{Mechanistic and targeted editing.}
Recent work pursues \emph{fine-grained} LLM unlearning by editing specific representations or parameter subspaces.
Representative approaches include RMU for safety-focused representation manipulation \citep{Li2024}, activation/subspace methods such as LUNAR, FALCON, UNLEARN, and Ethos \citep{Shen2025,Hu2025,Lizzo2024,Gao2024}, and parameter-selection strategies like selective pruning (LLM-Eraser) and influence reweighting (RapidUn) \citep{ZhangS2025,Zhao2025}.
While these methods clarify \emph{where} and \emph{how} to intervene, they typically rely on curated forget data or task vectors to specify \emph{what} to remove.
We use hallucination traces to localize hidden memorized content and apply sparse, utility-constrained gating to the responsible MLP pathways.

\textbf{Evaluating selectivity.}
MUSE measures verbatim reproduction and knowledge recall separately \citep{Shi2024MUSE}, and WMDP probes hazardous knowledge through multiple-choice questions \citep{Li2024}.
Open-ended prompts offer a complementary test because a single response can contain both material to forget and material to retain.
The SepS protocol used here combines forget and retain requests in a single prompt, testing whether an edit suppresses the target content while preserving other information in the same response.

\newpage
\section{Method}
\label{sec:method}

\subsection{Preliminary Study}
\label{sec:prelim}

\begin{wrapfigure}{R}{0.45\textwidth}  
\centering
\includegraphics[width=1\linewidth]{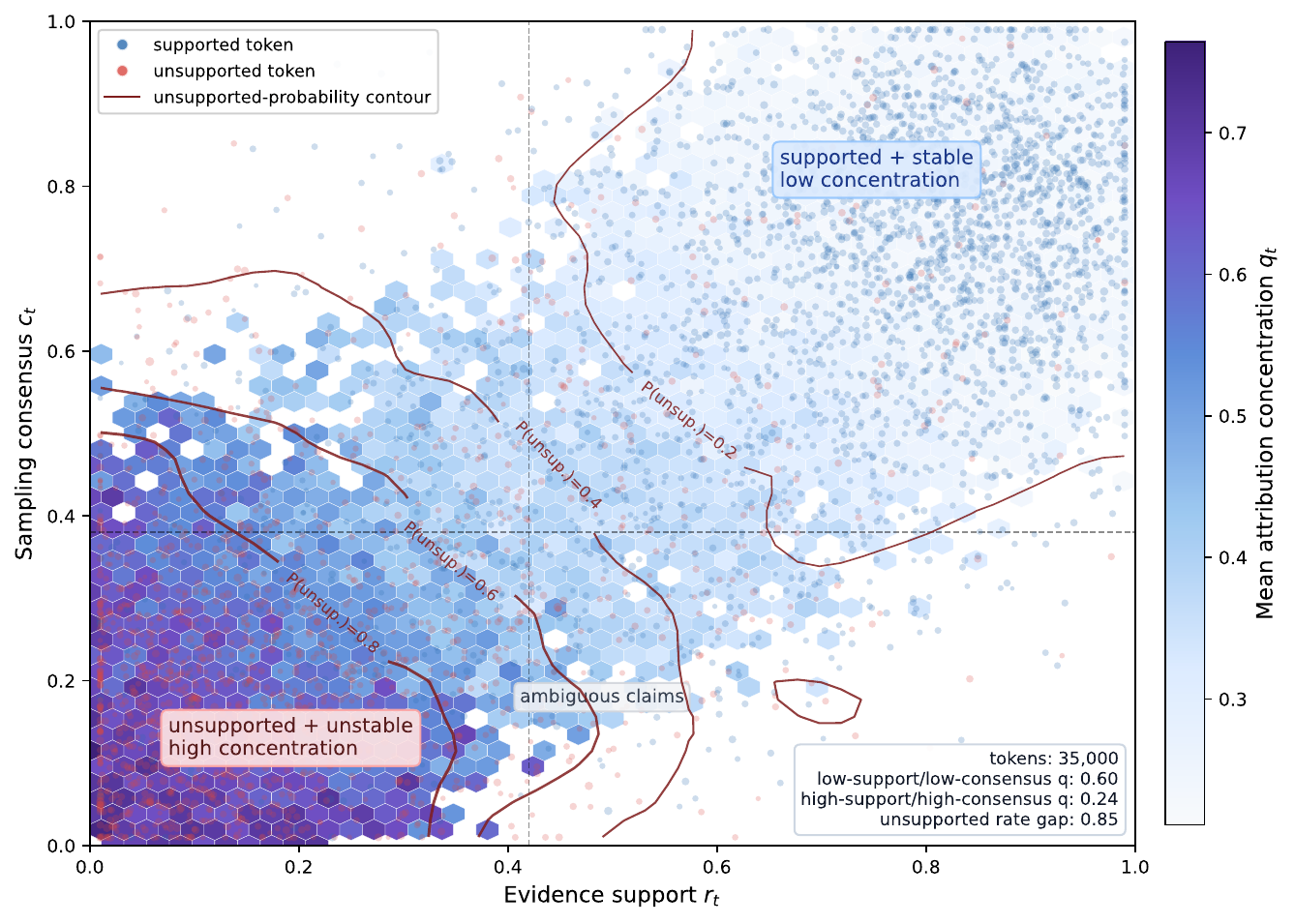}
\caption{
Unsupported and unstable generations concentrate in a high-concentration, high-confidence region.
}
\label{fig:prelim_phase_map}
\end{wrapfigure}

We run a preliminary diagnostic on a frozen aligned LLM.
For each open-ended prompt \(x\), we retrieve trusted evidence \(e(x)\), sample multiple responses, and decompose each response into token-level or short phrase-level units.
For every unit, we compute evidence support, cross-sample consensus, logit-margin confidence, and MLP attribution concentration.
We then plot the units in an evidence--consensus phase map: color shows attribution concentration, contours show the unsupported-token rate, and point size shows model confidence.
The phase map shows whether unsupported, low-consensus units also exhibit concentrated internal attribution.
It distinguishes uncertain from confident unsupported tokens. Confident unsupported tokens are more relevant to overfitted or memorized behavior.
The diagnostic measures support, stability, and attribution concentration before span or neuron selection, gating, and unlearning.
Detailed metric definitions are given in Appendix~\ref{app:prelim_protocol}.

\paragraph{Findings.} Figure~\ref{fig:prelim_phase_map} reveals three patterns.
First, unsupported tokens cluster where evidence support and sampling consensus are both low, indicating that external grounding and internal stability are complementary diagnostics (cf.\ structural and behavioral evidence fusion in cyber threat attribution \citealp{wang2026trident}).
Second, many unsupported tokens in this region have large logit margins.
Third, attribution concentration rises sharply in the same region, suggesting that hallucinated claims are often driven by sparse internal pathways.
These patterns suggest that hallucination traces contain a structured subset of externally unsupported and internally localized tokens.
This motivates using low-support, low-consensus, high-concentration tokens as diagnostic probes for the trace-guided unlearning procedure below.

\subsection{Overview}
\begin{figure}[h]
	\centering
	\includegraphics[width=\linewidth]{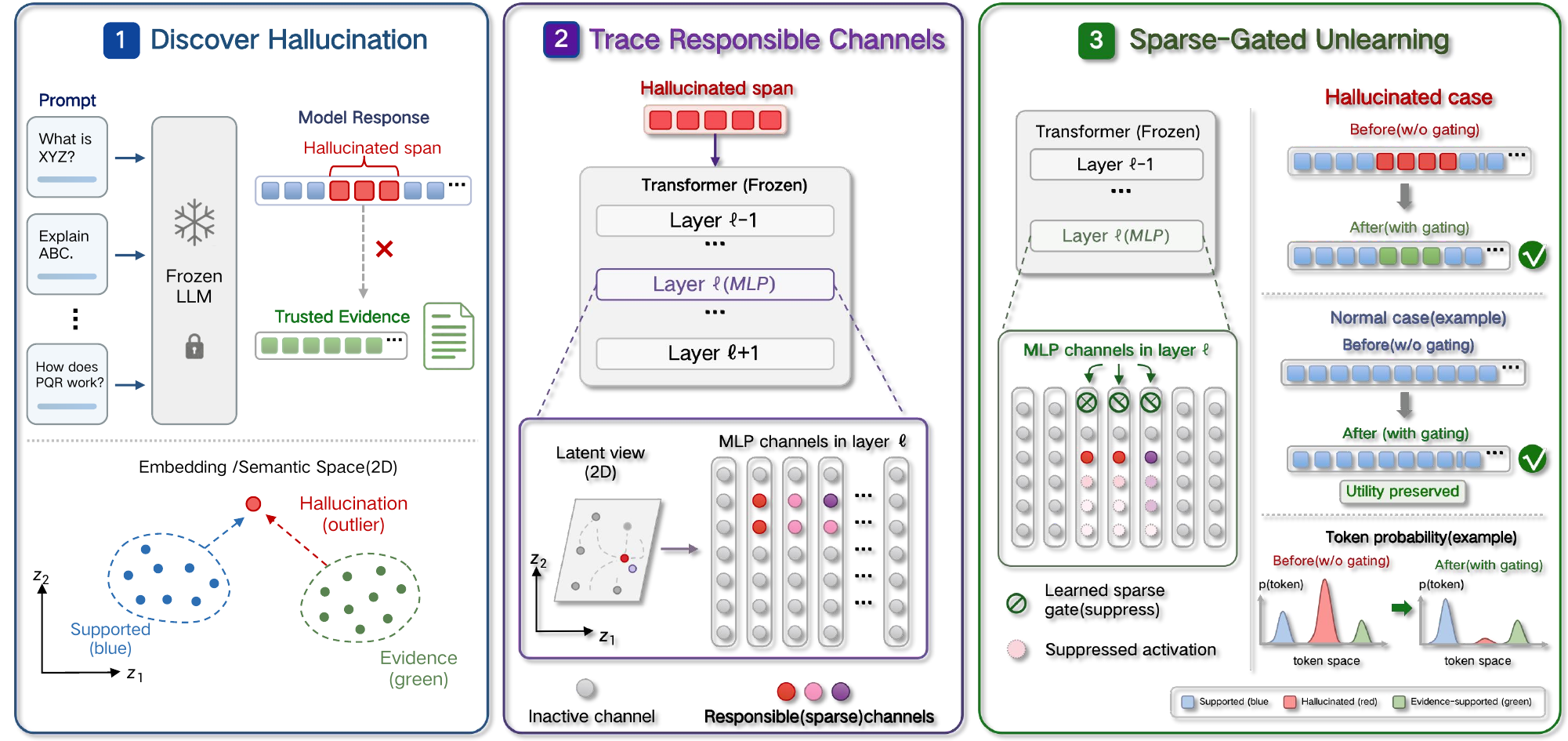}
	\caption{
\textbf{Overview of Hallucination-Guided Unlearning.}
HGU first discovers unsupported hallucinated spans by contrasting model responses with trusted evidence, then traces these spans to sparse high-responsibility MLP channels, and finally learns sparse gates to suppress hallucination-specific pathways while preserving normal evidence-supported behavior.
}
	\label{fig_pipeline}
\end{figure}

Hallucination-Guided Unlearning (HGU) uses \emph{diagnostic traces} from hallucinations to identify high-responsibility pathways and suppresses them with sparse trainable neuron gates on a frozen backbone. Starting from a base model $f_\theta$, HGU (i) selects hallucination-prone prompts and samples generations, (ii) localizes hallucinated spans via evidence support and stochastic-consensus, (iii) extracts neuron-level responsibility scores from MLP activations and their gradients, then (iv) trains sparse gates on the identified neurons to suppress hallucination-specific pathways while distilling the original behavior on a utility set. Figure~\ref{fig_pipeline} shows the pipeline.

\textbf{Notation and Model Internals.}
Let $f_\theta$ be an autoregressive LLM with vocabulary $\mathcal{V}$ and conditional distribution $p_\theta(y\mid x)=\prod_{t=1}^{T}p_\theta(y_t\mid x,y_{<t})$. We analyze a standard Transformer with $L$ layers. For layer $l$ and position $t$, let $A_{l,h,t,\cdot}$ be the attention distribution of head $h$ at query position $t$. Let $a_{l,t}\in\mathbb{R}^{d_{\mathrm{ff}}}$ denote the post-nonlinearity hidden activation of the MLP block at layer $l$ and position $t$ (the FFN hidden width is $d_{\mathrm{ff}}$). We compute all internal quantities using a single teacher-forced forward pass on the full sequence $(x,y)$.

\subsection{Hallucination Probing}
\label{sec:method_probe}
Hidden memorized content rarely exposes itself under typical prompts. HGU therefore constructs a probe set that increases the rate of hallucination, treating hallucinated generations as stress tests that surface overconfident, non-evidence-grounded continuations.

Given a query pool $\mathcal{Q}$, we retrieve supporting evidence for each prompt $x$ from a trusted corpus $\mathcal{K}$: $e(x)=\mathrm{Retrieve}(x;\mathcal{K})$. We then score prompts using a hallucination propensity that increases with model uncertainty and decreases with evidence support. Let
$\mathcal{H}_\theta(x)=-\sum_{w\in\mathcal{V}}p_\theta(w\mid x)\log p_\theta(w\mid x)$ be the next-token entropy at the first generation step, and let $\mathrm{Cov}(x,e(x))\in[0,1]$ be a deterministic evidence coverage score defined in Appendix~\ref{app:evidence}. We define
\begin{equation}
\pi(x)
\triangleq
\alpha_{\pi}\,\mathcal{H}_\theta(x)\;+\;\beta_{\pi}\,\big(1-\mathrm{Cov}(x,e(x))\big),
\label{eq:method_propensity}
\end{equation}
where $\alpha_{\pi}\ge 0$ and $\beta_{\pi}\ge 0$ are scalars. We select the top-$B$ prompts $\mathcal{X}=\mathrm{TopB}_{x\in\mathcal{Q}}\pi(x)$ and sample one completion per prompt via stochastic decoding, yielding probe pairs $\mathcal{D}_{\mathrm{H}}=\{(x,y)\}_{x\in\mathcal{X}}$.

\subsection{Span-Level Hallucination Localization}
\label{sec:method_span}
Unlearning requires localized targets. HGU assigns each generated token a hallucination weight that rises when the token lacks evidential support and remains unstable under stochastic-consensus at the same prefix.

Fix a probe pair $(x,y)\in\mathcal{D}_{\mathrm{H}}$. For each position $t$, we compute two scalar signals:

\textbf{Evidence support.}
Let $s_t$ be a short text span around $y_t$ (Appendix~\ref{app:spans}). A verifier $v_\psi$ maps $(s_t,e(x))$ to an entailment-like score in $[0,1]$.
We set $\phi_{\mathrm{ref}}(x,y,t)=v_\psi(s_t,e(x))$, where larger values indicate stronger support.

\textbf{Stochastic-consensus at fixed prefix.}
We run $K$ stochastic forward passes of $f_\theta$ at the same prefix $(x,y_{<t})$, yielding next-token distributions
$p_\theta^{(k)}(\cdot\mid x,y_{<t})$ for $k\in[K]$ (\S\ref{app:consensus}).
We define the consensus score as the mean probability assigned to the realized token $y_t$:
$\bar p_t \triangleq \frac{1}{K}\sum_{k=1}^{K}p_\theta^{(k)}(y_t\mid x,y_{<t})$,
and use disagreement $\phi_{\mathrm{con}}(x,y,t)\triangleq 1-\bar p_t$.

\textbf{Hallucination weight and span extraction.}
We combine the two signals into a single hallucination weight
\begin{equation}
h_t
\triangleq
\sigma\!\Big(
a\cdot\big(1-\phi_{\mathrm{ref}}(x,y,t)\big)
+
b\cdot\phi_{\mathrm{con}}(x,y,t)
-
c
\Big),
\label{eq:method_hall_weight}
\end{equation}
where $\sigma(u)=\frac{1}{1+e^{-u}}$ is the sigmoid, and $a\ge 0$, $b\ge 0$, $c\in\mathbb{R}$ are scalars.
We mark hallucinated indices by $\mathcal{T}_{\mathrm{H}}(x,y)=\{t: h_t\ge \tau\}$ with threshold $\tau\in(0,1)$ and extract maximal contiguous spans $\mathcal{S}(x,y)$ from $\mathcal{T}_{\mathrm{H}}(x,y)$.

\subsection{From Hallucination Traces to Responsible Neurons}
\label{sec:method_trace}
A hallucinated span is an output symptom. HGU turns this symptom into an internal target by identifying a sparse set of MLP neurons that (i) are highly active during hallucinated tokens and (ii) have large local influence on the log-probabilities of those tokens, with an additional weight that emphasizes low-entropy attention states associated with overconfident generation.

We evaluate internals on $(x,y)$ using teacher forcing. For each layer $l$ and hallucinated position $t\in\mathcal{T}_{\mathrm{H}}(x,y)$, define the token log-probability $\ell_t(\theta)\triangleq \log p_\theta(y_t\mid x,y_{<t})$ and the activation gradient
$g_{l,t}\triangleq \nabla_{a_{l,t}}\ell_t(\theta)\in\mathbb{R}^{d_{\mathrm{ff}}}$.
We compute an attention-entropy weight $w_{l,t}\ge 0$ from $A_{l,h,t,\cdot}$ (\S\ref{app:entropy}), where larger $w_{l,t}$ indicates lower attention entropy.

We score each MLP neuron $i\in[d_{\mathrm{ff}}]$ at layer $l$ by aggregating hallucination-weighted, entropy-weighted gradient--activation attribution:
\begin{equation}
I_{l,i}
\triangleq
\sum_{t\in \mathcal{T}_{\mathrm{H}}(x,y)}
h_t \cdot w_{l,t}\cdot \left| a_{l,t,i}\, g_{l,t,i}\right|,
\label{eq:method_neuron_score}
\end{equation}
where $a_{l,t,i}$ and $g_{l,t,i}$ denote the $i$-th coordinate of $a_{l,t}$ and $g_{l,t}$, respectively.
We select a fixed fraction of neurons per layer:
\begin{equation}
\mathcal{C}_l
\triangleq
\mathrm{Top}\_{\lceil\rho d_{\mathrm{ff}}\rceil}\!\left(\{I_{l,i}\}_{i=1}^{d_{\mathrm{ff}}}\right),
\qquad
\mathcal{C}\triangleq \bigcup_{l=1}^{L}\{(l,i): i\in\mathcal{C}_l\},
\label{eq:method_select}
\end{equation}
where $\rho\in(0,1)$ is the sparsity ratio. The set $\mathcal{C}$ is the hallucination memory channel set used by the unlearning intervention.

\subsection{Sparse-Gated Unlearning}
\label{sec:method_unlearn}
HGU trains a small set of neuron gates on $\mathcal{C}$ with the backbone frozen. The gates suppress hallucination-responsible neurons while distilling the original model behavior on a utility set, which stabilizes general performance.

We introduce trainable gates $\gamma_{l}\in[0,1]^{d_{\mathrm{ff}}}$ for each layer $l$, supported only on $\mathcal{C}_l$.
Let $a_{l,t}$ be the MLP hidden activation at layer $l$, position $t$. We apply gates by
\begin{equation}
\tilde a_{l,t}
\triangleq
a_{l,t}\odot(\mathbf{1}-\gamma_l),
\qquad
(\gamma_l)_i = 0\ \text{for all}\ i\notin\mathcal{C}_l,
\label{eq:method_gating}
\end{equation}
where $\odot$ is elementwise product and $\mathbf{1}$ is the all-ones vector. Replacing $a_{l,t}$ with $\tilde a_{l,t}$ at every layer yields an edited model distribution $p_{\theta,\gamma}(\cdot\mid x)$ with all backbone weights $\theta$ frozen.

\textbf{Training objective.}
We construct a utility set $\mathcal{D}_{\mathrm{U}}=\{(x^{u},y^{u})\}$ by sampling prompts and generating reference completions $y^{u}\sim p_\theta(\cdot\mid x^{u})$ using the base model once, then holding them fixed (\S\ref{app:utility}).
We minimize
\begin{equation}
\begin{aligned}
\min_{\gamma}\;&
\underbrace{
\mathbb{E}_{(x,y)\sim \mathcal{D}_{\mathrm{H}}}
\Bigg[
\sum_{t\in \mathcal{T}_{\mathrm{H}}(x,y)}
\log p_{\theta,\gamma}\!\big(y_t \mid x, y_{<t}\big)
\Bigg]
}_{\mathcal{L}_{\mathrm{forget}}(\gamma)}
\\
&\;+\;\lambda_{\mathrm{KL}}\,
\underbrace{
\mathbb{E}_{(x^{u},y^{u})\sim \mathcal{D}_{\mathrm{U}}}
\Bigg[
\sum_{t=1}^{|y^{u}|}
\Bigg(
\begin{aligned}
&\mathrm{KL}\!\Big(
p_{\theta}\!\big(\cdot \mid x^{u}, y^{u}_{<t}\big)
\\[-1pt]
&\qquad\qquad\big\|\;
p_{\theta,\gamma}\!\big(\cdot \mid x^{u}, y^{u}_{<t}\big)
\Big)
\end{aligned}
\Bigg)
\Bigg]
}_{\mathcal{L}_{\mathrm{keep}}(\gamma)}
\\
&\;+\;\lambda_{1}\|\gamma\|_{1}.
\end{aligned}
\label{eq:method_objective}
\end{equation}
where $\mathrm{KL}(\cdot\|\cdot)$ is the token-level Kullback--Leibler divergence, $\|\gamma\|_{1}=\sum_{l,i}|(\gamma_l)_i|$ promotes sparse edits, and $\lambda_{\mathrm{KL}}\ge 0$, $\lambda_1\ge 0$ are scalars.
Minimizing $\mathcal{L}_{\mathrm{forget}}$ reduces the probability of hallucinated tokens on hallucinated positions, while $\mathcal{L}_{\mathrm{keep}}$ distills the original next-token distributions to preserve general utility. We optimize Eq.~\eqref{eq:method_objective} with projected gradient updates that maintain $\gamma\in[0,1]^{Ld_{\mathrm{ff}}}$ (\S\ref{app:optimization}), then return the edited model $f_{\theta,\gamma}$.
Algorithm~\ref{alg:hgu} summarizes the complete procedure.

\begin{algorithm}[t]
\caption{Hallucination-Guided Unlearning (HGU)}
\label{alg:hgu}
\begin{algorithmic}[1]
\STATE \textbf{Input:} base model $f_\theta$, query pool $\mathcal{Q}$, corpus $\mathcal{K}$, budget $B$, sparsity $\rho$
\STATE Retrieve evidence $e(x)=\mathrm{Retrieve}(x;\mathcal{K})$ for each $x\in\mathcal{Q}$
\STATE Score prompts with $\pi(x)$ in Eq.~\eqref{eq:method_propensity} and select $\mathcal{X}=\mathrm{TopB}\,\pi(x)$
\STATE Sample one completion $y\sim p_\theta(\cdot\mid x)$ for each $x\in\mathcal{X}$ to form $\mathcal{D}_{\mathrm{H}}$
\FOR{each $(x,y)\in\mathcal{D}_{\mathrm{H}}$}
    \STATE Compute $\phi_{\mathrm{ref}}$ by Eq.~\eqref{eq:app_ref} and $\phi_{\mathrm{con}}$ by Eq.~\eqref{eq:app_consensus}
    \STATE Compute hallucination weights $h_t$ by Eq.~\eqref{eq:method_hall_weight} and extract $\mathcal{T}_{\mathrm{H}}(x,y)$
    \STATE Run teacher forcing on $(x,y)$ and compute neuron scores $I_{l,i}$ by Eq.~\eqref{eq:method_neuron_score}
\ENDFOR
\STATE Select neurons $\mathcal{C}$ by Eq.~\eqref{eq:method_select}
\STATE Build utility set $\mathcal{D}_{\mathrm{U}}$ as in Appendix~\ref{app:utility}
\STATE Train sparse gates $\gamma$ by minimizing Eq.~\eqref{eq:method_objective} with updates in Eq.~\eqref{eq:app_proj}
\STATE \textbf{Output:} edited model $f_{\theta,\gamma}$
\end{algorithmic}
\end{algorithm}

\section{Theory: Identifiability of Hallucination Memory Channels}
\label{sec:theory_ident}

\subsection{Trace Model and Estimator}
\label{sec:theory_ident_model}
We formalize when hallucination traces can \emph{recover the support} of the latent memory channels.
We analyze one layer (the analysis extends across layers by a union bound).
Fix a layer $l$ with MLP hidden width $d$ (we omit the subscript $l$).
For each independently sampled prompt $t\in[n]$, let $T_t$ contain its localized token positions and define the prompt-level trace
\begin{equation}
X_t\in\mathbb{R}_{\ge0}^{d},\qquad
(X_t)_i\triangleq\frac{1}{|T_t|}\sum_{r\in T_t}
\omega_{t,r}\,|a_{t,r,i}g_{t,r,i}|.
\end{equation}
Here $a_{t,r,i}$ is the post-nonlinearity MLP activation, $g_{t,r,i}$ is its exact local log-probability gradient, and $\omega_{t,r}\in[0,1]$ collects localization and attention-entropy weights; prompts with no selected tokens are omitted.
The empirical score is
\begin{equation}
\widehat I_i\triangleq\frac{1}{n}\sum_{t=1}^{n}(X_t)_i,\qquad i\in[d].
\label{eq:theory_score_def}
\end{equation}
The sufficient condition concerns support recovery in the idealized trace model. It uses a prompt-normalized analogue of Eq.~\eqref{eq:method_neuron_score} and assumes independence only across prompts.
Let $\mathcal{C}^\star\subset[d]$ be an idealized (unknown) high-responsibility support with $|\mathcal{C}^\star|=s$.
The estimator selects the top-$s$ coordinates:
$\widehat{\mathcal{C}} \triangleq \mathrm{Top}\_s(\{\widehat I_i\}_{i=1}^{d})$.

\begin{assumption}[Mean separation on hallucination traces]
\label{assump:sep}
Let $\mu_i\triangleq\mathbb{E}[(X_t)_i]$. There exists $\Delta>0$ such that
\begin{equation}
\min_{i\in\mathcal{C}^\star}\mu_i
-
\max_{j\notin\mathcal{C}^\star}\mu_j
\;\ge\;
\Delta.
\label{eq:theory_sep}
\end{equation}
\end{assumption}

\begin{assumption}[Sub-exponential trace noise]
\label{assump:subexp}
For every $i\in[d]$, the centered variables $(X_t)_i-\mu_i$ are independent across $t$ and sub-exponential with parameters $(\nu,b)$:
\begin{equation}
\mathbb{E}\big[\exp(\lambda((X_t)_i-\mu_i))\big]
\;\le\;
\exp\!\Big(\tfrac{\lambda^2\nu^2}{2}\Big)
\quad
\text{for all }\ |\lambda|\le 1/b.
\label{eq:theory_mgf}
\end{equation}
\end{assumption}

\subsection{Support Recovery and Sample Complexity}
\label{sec:theory_ident_main}
The next theorem states that hallucination traces identify the memory-channel support with high probability once the number of independent prompt-level traces $n$ exceeds a sharp Bernstein-type threshold.

\begin{theorem}[Support recovery from hallucination traces]
\label{thm:support_recovery}
Under Assumptions~\ref{assump:sep}--\ref{assump:subexp}, for any $\delta\in(0,1)$, if
\begin{equation}
n
\;\ge\;
\max\!\left\{
\frac{32\nu^2}{\Delta^2}\log\!\Big(\frac{2d}{\delta}\Big),
\ \ 
\frac{8b}{\Delta}\log\!\Big(\frac{2d}{\delta}\Big)
\right\},
\label{eq:theory_n_req}
\end{equation}
then $\widehat{\mathcal{C}}=\mathcal{C}^\star$ with probability at least $1-\delta$.
\end{theorem}

\textbf{Proof sketch.}
Let $\widehat I_i$ be defined in Eq.~\eqref{eq:theory_score_def}.
Step~1 proves a sub-exponential Bernstein bound for each coordinate:
\begin{equation}
\mathbb{P}\big(|\widehat I_i-\mu_i|\ge \varepsilon\big)
\le
2\exp\!\Big(-\tfrac{n}{2}\min\{\varepsilon^2/\nu^2,\ \varepsilon/b\}\Big),
\end{equation}
and Step~2 unions this over $d$ neurons with $\varepsilon=\Delta/4$, obtaining the event
$\mathcal{E}=\{\max_i|\widehat I_i-\mu_i|\le \Delta/4\}$ with probability $\ge 1-\delta$.
Step~3 shows that on $\mathcal{E}$, mean separation transfers to empirical separation:
\begin{equation}
\min_{i\in\mathcal{C}^\star}\widehat I_i
\;>\;
\max_{j\notin\mathcal{C}^\star}\widehat I_j,
\label{eq:theory_emp_sep}
\end{equation}
which forces $\mathrm{Top}_s(\widehat I)=\mathcal{C}^\star$. The full proof is in \S\ref{app:theory_ident}.
\qed

\begin{corollary}[Consistency]
\label{cor:consistency}
Under Assumptions~\ref{assump:sep}--\ref{assump:subexp}, $\mathbb{P}(\widehat{\mathcal{C}}=\mathcal{C}^\star)\to 1$ as $n\to\infty$.
\end{corollary}

\textbf{Remark (Near-tightness).}
\S\ref{app:lower_bound} gives an information-theoretic lower bound for a Gaussian trace model, showing that any procedure requires
$n=\Omega\big((\sigma^2/\Delta^2)\log(d/s)\big)$ samples for exact support recovery, matching Theorem~\ref{thm:support_recovery} up to constants and the $\log(d)$ vs.\ $\log(d/s)$ factor.

\begin{table}[t]
\centering
\small
\setlength{\tabcolsep}{4.5pt}
\renewcommand{\arraystretch}{1.15}
\caption{\textbf{TOFU main results (standard protocol).} We report MU, FE, and Avg for each split.
Best in \textbf{bold}, second best \underline{underlined}.}
\label{tab:main_tofu}
\begin{tabular}{lccc ccc ccc}
\toprule
& \multicolumn{3}{c}{\textbf{forget01}} & \multicolumn{3}{c}{\textbf{forget05}} & \multicolumn{3}{c}{\textbf{forget10}} \\
\cmidrule(lr){2-4}\cmidrule(lr){5-7}\cmidrule(lr){8-10}
\textbf{Method} & MU$\uparrow$ & FE$\uparrow$ & Avg$\uparrow$ & MU$\uparrow$ & FE$\uparrow$ & Avg$\uparrow$ & MU$\uparrow$ & FE$\uparrow$ & Avg$\uparrow$ \\
\midrule
GA+GD                    & 0.6696 & 0.5908 & 0.6302 & 0.0000 & \second{0.8772} & 0.4386 & 0.5592 & \best{0.9346} & \second{0.7469} \\
GA+KL                    & 0.6478 & 0.6030 & 0.6254 & 0.0000 & \best{0.9303} & 0.4651 & 0.0000 & \second{0.9020} & 0.4510 \\
NPO+GD                   & 0.6414 & 0.6109 & 0.6262 & \second{0.5465} & 0.6921 & 0.6193 & 0.5648 & 0.7668 & 0.6658 \\
NPO+KL                   & 0.6289 & 0.6065 & 0.6177 & 0.5231 & 0.7089 & 0.6160 & 0.4412 & 0.7680 & 0.6046 \\
DPO+GD                   & 0.7564 & 0.5335 & 0.6450 & 0.0000 & 0.8243 & 0.4122 & 0.0000 & 0.8041 & 0.4021 \\
DPO+KL                   & \second{0.7577} & 0.5383 & 0.6480 & 0.0000 & 0.8339 & 0.4169 & 0.0000 & 0.8421 & 0.4211 \\
IDK+GD                   & 0.6705 & \second{0.7697} & \second{0.7201} & 0.0000 & 0.7952 & 0.3976 & 0.0576 & 0.7603 & 0.4090 \\
\midrule
ULD (logit diff)         & 0.7012 & 0.6824 & 0.6918 & 0.5321 & 0.7356 & 0.6339 & 0.5813 & 0.7982 & 0.6898 \\
LLM-Eraser (pruning)     & 0.7125 & 0.6689 & 0.6907 & 0.5218 & 0.7483 & 0.6351 & 0.5927 & 0.8124 & 0.7026 \\
RapidUn (reweighting)    & 0.7241 & 0.6753 & 0.6997 & 0.5436 & 0.7429 & \second{0.6433} & \second{0.6015} & 0.8257 & 0.7136 \\
\midrule
\rowcolor{HGUrow} HGU (ours)      & \best{0.7932} & \best{0.7825} & \best{0.7879} & \best{0.6124} & 0.8317 & \best{0.7221} & \best{0.6483} & 0.8915 & \best{0.7699} \\
\bottomrule
\end{tabular}
\end{table}

\section{Experiments}
\label{sec:exp}

\subsection{Experimental Setup}
\label{sec:exp_setup}

\textbf{Datasets.}
We evaluate Hallucination-Guided Unlearning (HGU) on three widely adopted LLM-unlearning benchmarks that cover (i) entity-level memorization, (ii) copyrighted-corpus removal, and (iii) hazardous-knowledge suppression.
\textbf{TOFU} provides synthetic author profiles with Q\&A pairs and explicit \emph{forget} / \emph{retain} partitions for studying entity-level forgetting and collateral utility drift \citep{Maini2024}.
\textbf{MUSE} tests real-world corpus unlearning on \textbf{News} (BBC articles) and \textbf{Books} (Harry Potter) with a six-facet evaluation protocol that includes memorization, privacy leakage, and deployment constraints \citep{Shi2024MUSE}.
\textbf{WMDP} measures hazardous knowledge via multiple-choice questions in bio/cyber/chem security, serving as a safety-motivated unlearning benchmark \citep{Li2024}.
In addition, to test \emph{RLHF-aligned safety unlearning} under harmful prompts, we follow prior unlearning-for-safety setups and use \textbf{PKU-SafeRLHF} (forget) with \textbf{TruthfulQA} (utility) \citep{Liu2024}.

\textbf{Evaluation metrics.}
We use each benchmark’s standard metrics to compare with prior methods.
On \textbf{TOFU}, we report \emph{Forget Quality} and \emph{Model Utility} using the benchmark’s multi-metric protocol (generation similarity, likelihood-based measures, and truthfulness-based tests) across the Forget/Retain/Real-Author/World-Facts evaluation sets \citep{Maini2024}.
On \textbf{MUSE}, we report the six desiderata: \textbf{C1} no verbatim memorization, \textbf{C2} no knowledge memorization, \textbf{C3} no privacy leakage (membership inference), \textbf{C4} utility preservation, \textbf{C5} scalability to larger forget requests, and \textbf{C6} sustainability under sequential requests \citep{Shi2024MUSE}.
On \textbf{WMDP}, we report accuracy on WMDP-Bio/Cyber/Chem and standard utility probes (general knowledge and instruction-following) following the benchmark protocol \citep{Li2024}.
For \textbf{PKU-SafeRLHF}, we report harmfulness reduction on forget prompts together with utility preservation on normal prompts, matching the evaluation practice of safety-oriented unlearning \citep{Liu2024}.
We also report \textbf{Hallucination-Rate} on open-ended generations as the fraction of tokens/spans flagged as unsupported by reference evidence, which directly aligns with HGU’s design goal.

\textbf{Baselines.}
We compare against recent and representative LLM-unlearning methods spanning optimization-based objectives, logit/activation editing, and parameter selection.
\textbf{Optimization-based} baselines include Gradient Ascent and its retain-regularized variants (Gradient Difference / retain-side KL), as well as \textbf{NPO} \citep{Maini2024,ZhangR2024,Shi2024MUSE}.
\textbf{Logit-space} unlearning is represented by \textbf{ULD} \citep{Ji2024}.
\textbf{Representation/activation} approaches include \textbf{RMU} (WMDP), \textbf{LUNAR}, \textbf{UNLEARN}, \textbf{Ethos}, and \textbf{FALCON} \citep{Li2024,Shen2025,Lizzo2024,Gao2024,Hu2025}.
\textbf{Parameter selection} baselines include selective pruning and influence-driven reweighting \citep{ZhangS2025,Zhao2025}.
\textbf{Prompt-/decoding-time} controls (e.g., SPUL, in-context unlearning, DExperts, and pretend-to-forget prompting) are included as behavioral baselines \citep{Bhaila2025,Pawelczyk2023,Liu2021,Takashiro2025}.
Refer to \S\ref{app:exp_setup} for implementation details.

\begin{table}[t]
\centering
\small
\setlength{\tabcolsep}{4.2pt}
\renewcommand{\arraystretch}{1.12}
\caption{\textbf{TOFU under the SepS protocol (mixed-prompt evaluation).}
We report MU/FE/SepS and their harmonic mean (H-Avg.).}
\label{tab:tofu_seps}
\resizebox{\textwidth}{!}{%
\begin{tabular}{lcccc cccc cccc}
\toprule
& \multicolumn{4}{c}{\textbf{forget01}} & \multicolumn{4}{c}{\textbf{forget05}} & \multicolumn{4}{c}{\textbf{forget10}} \\
\cmidrule(lr){2-5}\cmidrule(lr){6-9}\cmidrule(lr){10-13}
\textbf{Method} & MU$\uparrow$ & FE$\uparrow$ & SepS$\uparrow$ & H-Avg.$\uparrow$ &
MU$\uparrow$ & FE$\uparrow$ & SepS$\uparrow$ & H-Avg.$\uparrow$ &
MU$\uparrow$ & FE$\uparrow$ & SepS$\uparrow$ & H-Avg.$\uparrow$ \\
\midrule
\multicolumn{13}{l}{\textit{Untargeted unlearning}}\\
GA+GD   & 0.7043 & 0.6233 & 0.0225 & 0.0631 & 0.1061 & 0.9075 & 0.0063 & 0.0177 & 0.5263 & 0.9271 & 0.0065 & 0.0192 \\
GA+KL   & 0.7109 & 0.6189 & 0.0284 & 0.0784 & 0.0000 & 0.8980 & 0.0001 & 0.0000 & 0.0000 & 0.9438 & 0.0007 & 0.0000 \\
NPO+GD  & \second{0.7196} & 0.6371 & 0.0347 & 0.0945 & 0.5469 & 0.8300 & 0.0331 & 0.0903 & 0.4515 & 0.8045 & 0.0675 & 0.1642 \\
NPO+KL  & 0.7150 & 0.6366 & 0.0298 & 0.0822 & 0.3657 & 0.8172 & 0.0647 & 0.1546 & 0.2111 & 0.8418 & 0.0070 & 0.0203 \\
ME+GD   & 0.7165 & \second{0.9694} & 0.0395 & 0.1081 & \second{0.6769} & \second{0.9703} & 0.0309 & 0.0860 & \second{0.6966} & \second{0.9660} & 0.0376 & 0.1031 \\
\midrule
\multicolumn{13}{l}{\textit{Targeted unlearning}}\\
DPO+GD  & 0.4534 & 0.7782 & 0.1584 & 0.3059 & 0.0140 & 0.7910 & 0.0174 & 0.0230 & 0.2552 & 0.7429 & 0.0914 & 0.1851 \\
DPO+KL  & 0.4389 & 0.7826 & 0.1570 & 0.3022 & 0.0000 & 0.8350 & 0.0000 & 0.0000 & 0.0131 & 0.8192 & 0.0002 & 0.0004 \\
IDK+GD  & 0.6123 & 0.7063 & 0.2005 & 0.3733 & 0.0237 & 0.7529 & 0.0019 & 0.0052 & 0.5045 & 0.7227 & 0.0733 & 0.1763 \\
IDK+KL  & 0.6099 & 0.7131 & 0.2004 & 0.3734 & 0.0000 & 0.8068 & 0.0000 & 0.0000 & 0.0446 & 0.7674 & 0.0006 & 0.0018 \\
IDK+AP  & 0.6810 & 0.7744 & \second{0.2787} & \second{0.4726} & 0.6600 & 0.7293 & \second{0.1500} & \second{0.3140} & 0.6129 & 0.7318 & \second{0.1490} & \second{0.3090} \\
\midrule
\rowcolor{HGUrow} HGU (ours) & \best{0.7324} & \best{0.9812} & \best{0.3254} & \best{0.5497} & \best{0.7215} & \best{0.9786} & \best{0.2253} & \best{0.4382} & \best{0.7238} & \best{0.9753} & \best{0.2528} & \best{0.4715} \\
\bottomrule
\end{tabular}%
}
\end{table}

\begin{table}[t]
\centering
\small
\setlength{\tabcolsep}{5.8pt}
\renewcommand{\arraystretch}{1.15}
\caption{\textbf{WMDP auxiliary benchmark (ZEPHYR-7B).}
Lower is better for \textbf{Forget}, higher is better for \textbf{Retain} and \textbf{MMLU}.}
\label{tab:main_wmdp_aux}
\resizebox{\textwidth}{!}{%
\begin{tabular}{lccc ccc ccc}
\toprule
& \multicolumn{3}{c}{\textbf{Physics}} & \multicolumn{3}{c}{\textbf{Law}} & \multicolumn{3}{c}{\textbf{Economics}} \\
\cmidrule(lr){2-4}\cmidrule(lr){5-7}\cmidrule(lr){8-10}
\textbf{Method} & Forget$\downarrow$ & Retain$\uparrow$ & MMLU$\uparrow$ &
Forget$\downarrow$ & Retain$\uparrow$ & MMLU$\uparrow$ &
Forget$\downarrow$ & Retain$\uparrow$ & MMLU$\uparrow$ \\
\midrule
Base (no unlearning) & 38.8 & 34.6 & \second{58.6} & 56.7 & \best{71.3} & \best{58.6} & 60.2 & 45.6 & \best{58.6} \\
RMU                 & 27.0 & 29.2 & 57.1 & 27.8 & 37.0 & 54.5 & 27.3 & 41.2 & 55.0 \\
\midrule
FALCON              & 24.5 & 30.1 & 57.3 & 25.2 & 38.5 & 54.8 & 24.8 & 42.8 & 55.2 \\
LUNAR               & 22.8 & 31.5 & 57.5 & 23.6 & 40.2 & 55.1 & 22.5 & 44.3 & 55.5 \\
UNLEARN             & 21.3 & 32.8 & 57.8 & 22.1 & 41.7 & 55.4 & 20.9 & 45.9 & 55.9 \\
Ethos               & 19.7 & 33.5 & 58.0 & 20.4 & 43.2 & 55.7 & 19.2 & 47.3 & 56.2 \\
LLM-Eraser          & 18.5 & 34.2 & 58.2 & 19.1 & 44.6 & 56.0 & 18.0 & 48.5 & 56.5 \\
RapidUn             & \second{17.2} & \second{35.8} & 58.4 & \second{17.9} & 46.3 & 56.3 & \second{16.8} & \second{50.1} & 56.8 \\
\midrule
\rowcolor{HGUrow} HGU (ours) & \best{14.6} & \best{37.2} & \best{58.9} & \best{15.3} & \second{48.9} & \second{56.8} & \best{14.2} & \best{52.7} & \second{57.2} \\
\bottomrule
\end{tabular}%
}
\end{table}

\subsection{Main Results}
\label{sec:main_results}

\textbf{Results on TOFU.}
Table~\ref{tab:main_tofu} summarizes the standard TOFU protocol using Model Utility (MU), Forget Efficacy (FE), and their average (Avg).
We report prior methods’ numbers when available and evaluate HGU under the same backbone and decoding setup.
HGU achieves the best overall MU--FE trade-off across \texttt{forget01/forget05/forget10}, improving Avg consistently over strong optimization-, logit-, and pruning-based baselines.
We further validate robustness under mixed forget/retain prompts using the SepS protocol in Table~\ref{tab:tofu_seps}, where HGU yields the highest harmonic mean (H-Avg.). Appendix~\ref{app.abl} reports ablations.

\textbf{Results on WMDP auxiliary unlearning (MMLU subsets).}
Table~\ref{tab:main_wmdp_aux} reports ZEPHYR-7B results on the WMDP MMLU-subset auxiliary task, where the objective is to lower \textbf{Forget} accuracy while preserving \textbf{Retain} and overall \textbf{MMLU} performance.
Across Physics, Law, and Economics, HGU achieves the strongest forgetting and the highest Retain and MMLU scores among edited models, indicating a favorable safety--utility trade-off.

\subsection{Analysis}
\label{sec:ablation}

\begin{figure}[t]
\centering
\includegraphics[width=0.9\linewidth]{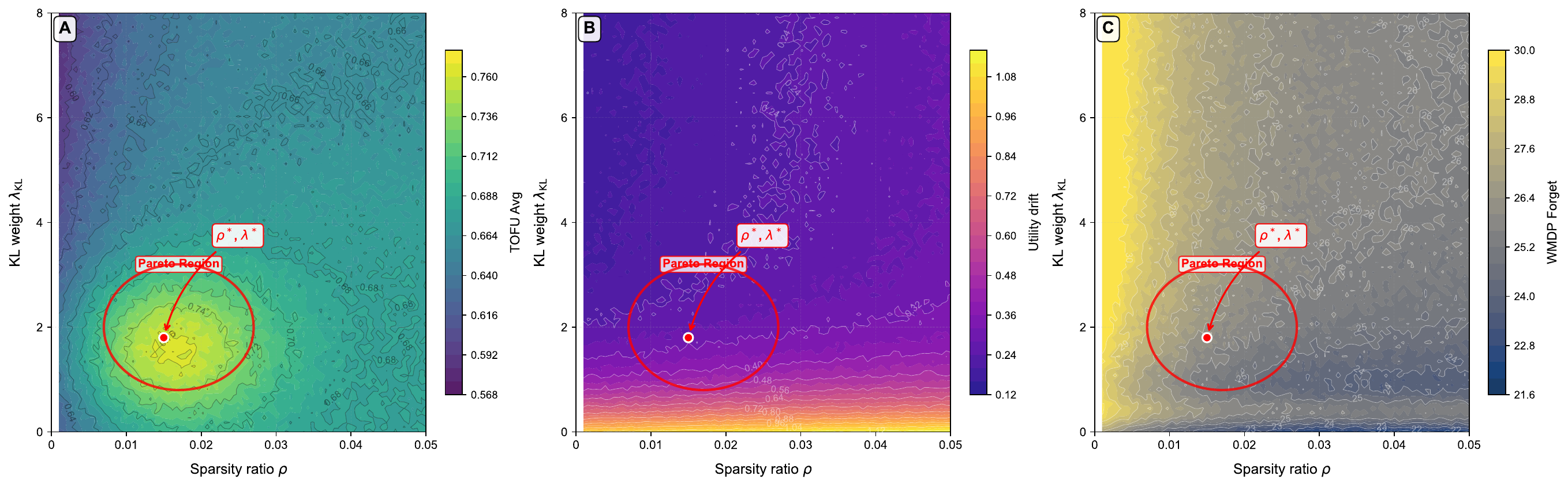}
\vspace{-2mm}
\caption{\textbf{Compression--Drift--Utility landscapes over $(\rho,\lambda_{\mathrm{KL}})$.}
\textbf{Left:} filled contours of TOFU Avg$\uparrow$ (higher is better).
\textbf{Middle:} filled contours of utility drift (token-level KL)$\downarrow$ (lower is better).
\textbf{Right:} filled contours of WMDP Forget$\downarrow$ (lower is better).
The dashed ellipse highlights a Pareto ``sweet spot'' where TOFU remains high while drift stays low and WMDP forgetting improves.
The arrow marks a recommended operating point (moderate sparsity + moderate $\lambda_{\mathrm{KL}}$) that balances all three criteria.}
\label{fig:pareto_contours}
\vspace{-3mm}
\end{figure}

\textbf{Compression--Drift--Utility.}
We visualize how HGU’s two primary control knobs---the neuron sparsity ratio $\rho$ (compression) and the keep-KL weight $\lambda_{\mathrm{KL}}$ (utility preservation)---jointly shape the trade-off among \emph{(i) forgetting quality}, \emph{(ii) utility drift}, and \emph{(iii) safety forgetting on WMDP}.
We test whether a \emph{Pareto region} combines strong forgetting with low utility drift under moderate compression.
Figure~\ref{fig:pareto_contours} shows a structured trade-off.
At very small $\lambda_{\mathrm{KL}}$, forgetting is stronger and drift spikes. At very large $\lambda_{\mathrm{KL}}$, utility is preserved and WMDP forgetting is limited.
The left and middle panels share a broad region of high TOFU performance and low drift at moderate compression, indicating that trace-guided sparsity can preserve utility.

\begin{wrapfigure}{r}{0.59\textwidth}
\centering
\includegraphics[width=\linewidth]{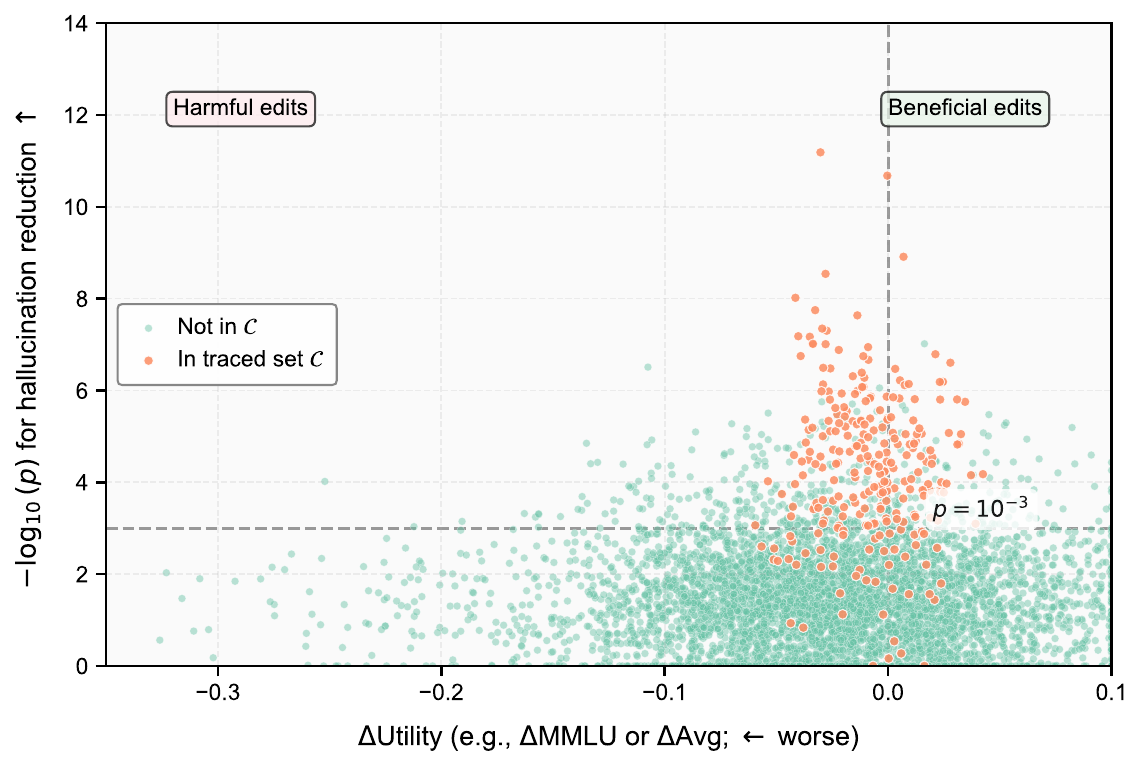}
\caption{\textbf{Channel intervention sensitivity.}
Each point is one channel intervention.
Channels in the traced set $\mathcal{C}$ cluster in the \emph{upper region near $\Delta\mathrm{Utility}\approx 0$}, indicating large hallucination reductions with minimal utility change, whereas non-traced channels either have small effects or require larger utility loss to achieve comparable reductions.}
\label{fig:volcano}
\end{wrapfigure}
\textbf{Channel Intervention Sensitivity.}
We test whether the channels selected by hallucination traces are \emph{causally} responsible for undesirable behavior.
For each candidate channel, we apply a small isolated intervention (masking or gating) and measure (i) the resulting change in utility and (ii) the statistical significance of hallucination reduction.
If the traced set $\mathcal{C}$ captures overfitted memory pathways, interventions on $\mathcal{C}$ should substantially reduce hallucinations with little utility change.
Figure~\ref{fig:volcano} shows that channels in $\mathcal{C}$ dominate the high-significance region while remaining close to $\Delta\mathrm{Utility}=0$, supporting a causal interpretation of the trace scores.
Channels outside $\mathcal{C}$ rarely achieve comparable significance at similarly low utility cost.
This pattern suggests that hallucination traces identify a compact set of intervention points that offer a favorable forgetting--utility trade-off, consistent with our identifiability analysis in \S\ref{sec:theory_ident}.

\newpage
\section{Conclusion}
\label{sec:conclusion}
We introduced \textsc{Hallucination-Guided Unlearning} (HGU), which uses hallucination traces to localize targets and perform sparse, utility-constrained edits to internal memory pathways.
Attribution for hallucinated tokens concentrates in a small set of neurons. The results suggest that precise localization improves the forgetting--utility trade-off.
On TOFU, HGU achieves the strongest balance between forgetting and utility among the evaluated methods under both standard and mixed forget/retain prompts.
On the WMDP auxiliary task, HGU reduces target accuracy while retaining higher utility than the other edited models.
Together, these results illustrate how unsupported generations can help identify candidate targets for selective editing when evidence or policy can guide the intervention.

\begin{ack}
\textbf{Funding.} The authors received no third-party funding or third-party support for this work.

\textbf{Competing interests.} The authors declare no competing financial interests or relevant financial relationships requiring disclosure.
\end{ack}

\bibliographystyle{plainnat}
\bibliography{references}

\begin{thebibliography}{36}
\providecommand{\natexlab}[1]{#1}
\providecommand{\url}[1]{\texttt{#1}}
\expandafter\ifx\csname urlstyle\endcsname\relax
  \providecommand{\doi}[1]{doi: #1}\else
  \providecommand{\doi}{doi: \begingroup \urlstyle{rm}\Url}\fi

\bibitem[Bhaila et~al.(2025)Bhaila, Van, and Wu]{Bhaila2025}
Karuna Bhaila, Minh-Hao Van, and Xintao Wu.
\newblock Soft prompting for unlearning in large language models.
\newblock In \emph{Proceedings of the 2025 Conference of the Nations of the
  Americas Chapter of the Association for Computational Linguistics: Human
  Language Technologies (Volume 1: Long Papers)}, pages 4046--4056, 2025.

\bibitem[Bourtoule et~al.(2021)Bourtoule, Chandrasekaran, Choquette-Choo, Jia,
  Travers, Zhang, Lie, and Papernot]{Bourtoule2021}
Lucas Bourtoule, Varun Chandrasekaran, Christopher~A Choquette-Choo, Hengrui
  Jia, Adelin Travers, Baiwu Zhang, David Lie, and Nicolas Papernot.
\newblock Machine unlearning.
\newblock In \emph{2021 IEEE symposium on security and privacy (SP)}, pages
  141--159. IEEE, 2021.

\bibitem[Chen et~al.(2025)Chen, Jiang, Liu, Gupta, Li, Zhao, and
  Wang]{Chen2025PhysGen3D}
Boyuan Chen, Hanxiao Jiang, Shaowei Liu, Saurabh Gupta, Yunzhu Li, Hao Zhao,
  and Shenlong Wang.
\newblock {PhysGen3D}: Crafting a miniature interactive world from a single
  image.
\newblock In \emph{Proceedings of the IEEE/CVF Conference on Computer Vision
  and Pattern Recognition (CVPR)}, pages 6178--6189, 2025.
\newblock \doi{10.1109/CVPR52734.2025.00579}.

\bibitem[Gao et~al.(2024)Gao, Niu, Tang, Avestimehr, and Annavaram]{Gao2024}
Lei Gao, Yue Niu, Tingting Tang, Salman Avestimehr, and Murali Annavaram.
\newblock Ethos: Rectifying language models in orthogonal parameter space.
\newblock In \emph{Findings of the Association for Computational Linguistics:
  NAACL 2024}, pages 2054--2068, 2024.

\bibitem[Hu et~al.(2025)Hu, Huang, Yin, Ruan, Cheng, Dong, and Huang]{Hu2025}
Jinwei Hu, Zhenglin Huang, Xiangyu Yin, Wenjie Ruan, Guangliang Cheng, Yi~Dong,
  and Xiaowei Huang.
\newblock Falcon: Fine-grained activation manipulation by contrastive
  orthogonal unalignment for large language model.
\newblock \emph{arXiv preprint arXiv:2502.01472}, 2025.

\bibitem[Hu(2026)]{hu2026research}
Saisai Hu.
\newblock Research on security enhancement methods for adversarial robust large
  language model intelligent agents for medical decision-making tasks.
\newblock In \emph{2026 7th International Seminar on Artificial Intelligence,
  Networking and Information Technology (AINIT)}, pages 599--603. IEEE, 2026.

\bibitem[Ji et~al.(2024)Ji, Liu, Zhang, Liu, Kompella, Liu, and Chang]{Ji2024}
Jiabao Ji, Yujian Liu, Yang Zhang, Gaowen Liu, Ramana~R Kompella, Sijia Liu,
  and Shiyu Chang.
\newblock Reversing the forget-retain objectives: An efficient llm unlearning
  framework from logit difference.
\newblock \emph{Advances in Neural Information Processing Systems},
  37:\penalty0 12581--12611, 2024.

\bibitem[Li et~al.(2024)Li, Pan, Gopal, Yue, Berrios, Gatti, Li, Dombrowski,
  Goel, Phan, et~al.]{Li2024}
Nathaniel Li, Alexander Pan, Anjali Gopal, Summer Yue, Daniel Berrios, Alice
  Gatti, Justin~D Li, Ann-Kathrin Dombrowski, Shashwat Goel, Long Phan, et~al.
\newblock The wmdp benchmark: Measuring and reducing malicious use with
  unlearning.
\newblock \emph{arXiv preprint arXiv:2403.03218}, 2024.

\bibitem[Li et~al.(2023)Li, Chen, Zhao, Gong, Zhou, Rossano, and
  Zhu]{Li2023TouchLine}
Yang Li, Xiaoxue Chen, Hao Zhao, Jiangtao Gong, Guyue Zhou, Federico Rossano,
  and Yixin Zhu.
\newblock Understanding embodied reference with {Touch-Line Transformer}.
\newblock In \emph{International Conference on Learning Representations
  (ICLR)}, 2023.
\newblock URL
  \url{https://openreview.net/pdf/f05337a1a545b0bea49b8f29bbfa4c42519f1b18.pdf}.

\bibitem[Liu et~al.(2021)Liu, Sap, Lu, Swayamdipta, Bhagavatula, Smith, and
  Choi]{Liu2021}
Alisa Liu, Maarten Sap, Ximing Lu, Swabha Swayamdipta, Chandra Bhagavatula,
  Noah~A Smith, and Yejin Choi.
\newblock Dexperts: Decoding-time controlled text generation with experts and
  anti-experts.
\newblock In \emph{Proceedings of the 59th Annual Meeting of the Association
  for Computational Linguistics and the 11th International Joint Conference on
  Natural Language Processing (Volume 1: Long Papers)}, 2021.

\bibitem[Liu et~al.(2026)Liu, Yang, Qin, and Zhang]{liu2026interpretivelockin}
Jingjing Liu, Dishu Yang, Hongbo Qin, and Hansong Zhang.
\newblock Passing tests, missing semantics: Evidence of interpretive lock-in in
  plan-based {DataFrame} code generation.
\newblock arXiv preprint arXiv:2610.15797, 2026.
\newblock URL \url{https://arxiv.org/abs/2610.15797}.

\bibitem[Liu et~al.(2024)Liu, Dou, Tan, Tian, and Jiang]{Liu2024}
Zheyuan Liu, Guangyao Dou, Zhaoxuan Tan, Yijun Tian, and Meng Jiang.
\newblock Towards safer large language models through machine unlearning.
\newblock In \emph{Findings of the Association for Computational Linguistics
  ACL 2024}, pages 1817--1829, 2024.

\bibitem[Lizzo and Heck(2025)]{Lizzo2024}
Tyler Lizzo and Larry Heck.
\newblock Unlearn efficient removal of knowledge in large language models.
\newblock In \emph{Findings of the Association for Computational Linguistics:
  NAACL 2025}, pages 7257--7268, 2025.

\bibitem[Ma et~al.(2026{\natexlab{a}})Ma, Hu, Zhu, Lin, Sun, Weng, Li, and
  Zhang]{ma2026musk}
Mingjun Ma, Saisai Hu, Tiantian Zhu, Yitian Lin, Jiachen Sun, Zhengqiu Weng,
  Dehao Li, and Yajie Zhang.
\newblock {MuSK}: Multi-scale knowledge learning for provenance-graph anomaly
  detection.
\newblock \emph{Computer Networks}, 289:\penalty0 112728, 2026{\natexlab{a}}.
\newblock \doi{10.1016/j.comnet.2026.112728}.

\bibitem[Ma et~al.(2026{\natexlab{b}})Ma, Luo, Zhangji, Xiao, Gao, Wang, Wu,
  Li, Wei, and Zhang]{Ma2026HiddenAPIs}
SiYuan Ma, Yiqin Luo, Zhangji, Canran Xiao, Albert Gao, Wei Wang, Qiwei Wu,
  Xinran Li, Jinfeng Wei, and Qixin Zhang.
\newblock Hidden {APIs} in language models: Discovering reusable causal
  interfaces from forked futures.
\newblock \emph{arXiv preprint arXiv:2607.27617}, 2026{\natexlab{b}}.
\newblock URL \url{https://arxiv.org/abs/2607.27617v2}.

\bibitem[Maini et~al.(2024)Maini, Feng, Schwarzschild, Lipton, and
  Kolter]{Maini2024}
Pratyush Maini, Zhili Feng, Avi Schwarzschild, Zachary~Chase Lipton, and J~Zico
  Kolter.
\newblock Tofu: A task of fictitious unlearning for llms.
\newblock In \emph{First Conference on Language Modeling}, 2024.

\bibitem[Pan et~al.(2025)Pan, Zhang, Zheng, Li, Cheng, and Zhao]{Pan2025}
Zibin Pan, Shuwen Zhang, Yuesheng Zheng, Chi Li, Yuheng Cheng, and Junhua Zhao.
\newblock Multi-objective large language model unlearning.
\newblock In \emph{ICASSP 2025-2025 IEEE International Conference on Acoustics,
  Speech and Signal Processing (ICASSP)}, pages 1--5. IEEE, 2025.

\bibitem[Pawelczyk et~al.(2024)Pawelczyk, Neel, and Lakkaraju]{Pawelczyk2023}
Martin Pawelczyk, Seth Neel, and Himabindu Lakkaraju.
\newblock In-context unlearning: Language models as few-shot unlearners.
\newblock In \emph{International Conference on Machine Learning}, pages
  40034--40050. PMLR, 2024.

\bibitem[Shao(2026{\natexlab{a}})]{shao2026interpretable}
Wanru Shao.
\newblock Interpretable ensemble learning for network traffic anomaly
  detection: A {SHAP}-based explainable {AI} framework for embedded systems
  security.
\newblock In \emph{2026 14th International Symposium on Digital Forensics and
  Security (ISDFS)}, pages 1--4. IEEE, 2026{\natexlab{a}}.
\newblock \doi{10.1109/ISDFS69419.2026.11459069}.
\newblock URL \url{https://doi.org/10.1109/ISDFS69419.2026.11459069}.

\bibitem[Shao(2026{\natexlab{b}})]{shao2026policy}
Wanru Shao.
\newblock Policy-driven vulnerability risk quantification framework for
  large-scale cloud infrastructure data security.
\newblock In \emph{Proceedings of the 2026 5th International Conference on
  Cyber Security, Artificial Intelligence and Digital Economy}, pages 40--45,
  New York, NY, USA, 2026{\natexlab{b}}. Association for Computing Machinery.
\newblock \doi{10.1145/3819836.3819842}.
\newblock URL \url{https://doi.org/10.1145/3819836.3819842}.

\bibitem[Shen et~al.(2025)Shen, Qiu, Kurmanji, Iacob, Sani, Chen, Cancedda, and
  Lane]{Shen2025}
William~F Shen, Xinchi Qiu, Meghdad Kurmanji, Alex Iacob, Lorenzo Sani, Yihong
  Chen, Nicola Cancedda, and Nicholas~D Lane.
\newblock Lunar: Llm unlearning via neural activation redirection.
\newblock \emph{arXiv preprint arXiv:2502.07218}, 2025.

\bibitem[Shi et~al.(2025)Shi, Lee, Huang, Malladi, Zhao, Holtzman, Liu,
  Zettlemoyer, Smith, and Zhang]{Shi2024MUSE}
Weijia Shi, Jaechan Lee, Yangsibo Huang, Sadhika Malladi, Jieyu Zhao, Ari
  Holtzman, Daogao Liu, Luke Zettlemoyer, Noah~A Smith, and Chiyuan Zhang.
\newblock Muse: Machine unlearning six-way evaluation for language models.
\newblock In \emph{The Thirteenth International Conference on Learning
  Representations}, 2025.

\bibitem[Takashiro et~al.(2025)Takashiro, Kojima, Gambardella, Cao, Iwasawa,
  and Matsuo]{Takashiro2025}
Shota Takashiro, Takeshi Kojima, Andrew Gambardella, Qi~Cao, Yusuke Iwasawa,
  and Yutaka Matsuo.
\newblock Answer when needed, forget when not: Language models pretend to
  forget via in-context knowledge unlearning.
\newblock In \emph{Findings of the Association for Computational Linguistics:
  ACL 2025}, pages 24872--24885, 2025.

\bibitem[Wang and Saxena(2026)]{wang2026tradingagentsmarketcrashers}
Mengxiao Wang and Nitesh Saxena.
\newblock Trading agents or market crashers? dissecting robustness and security
  failures in academic financial {LLM} trading schemes, 2026.
\newblock URL \url{https://arxiv.org/abs/2609.19705}.

\bibitem[Wang et~al.(2026)Wang, Hu, Zhu, Cheng, Weng, Chen, Zheng, and
  Zhang]{wang2026trident}
Zijun Wang, Saisai Hu, Tiantian Zhu, Wenrui Cheng, Zhengqiu Weng, Haiting Chen,
  Xiangyang Zheng, and Suyu Zhang.
\newblock {Trident}: Dual-stream {APT} attribution over heterogeneous threat
  knowledge graphs.
\newblock \emph{Computers \& Security}, 171:\penalty0 105094, 2026.
\newblock \doi{10.1016/j.cose.2026.105094}.

\bibitem[Xiao et~al.(2026{\natexlab{a}})Xiao, Chen, and Hou]{Xiao2026IDFT}
Canran Xiao, Qianyu Chen, and Liwei Hou.
\newblock Influence-disentangled federated training: Learning models that are
  easy to unlearn.
\newblock In \emph{Proceedings of the 43rd International Conference on Machine
  Learning}, volume 306 of \emph{Proceedings of Machine Learning Research},
  pages 138629--138649. PMLR, 2026{\natexlab{a}}.
\newblock URL \url{https://proceedings.mlr.press/v306/xiao26b.html}.

\bibitem[Xiao et~al.(2026{\natexlab{b}})Xiao, Gudepu, and
  Fumagalli]{xiao2026cora}
Linqi Xiao, Venkateswarlu Gudepu, and Andrea Fumagalli.
\newblock {CORA}: A conversational {AI}-agent for intelligent automation of
  {OpenROADM} multi-vendor optical transport networks.
\newblock In \emph{Applications of Machine Learning 2026}, volume 14218 of
  \emph{Proceedings of {SPIE}}, page 142180Y. SPIE, 2026{\natexlab{b}}.
\newblock \doi{10.1117/12.3113861}.

\bibitem[Xiao et~al.(2026{\natexlab{c}})Xiao, Latha, Gudepu, and
  Fumagalli]{xiao2026yang}
Linqi Xiao, Aparaajitha~Gomathinayakam Latha, Venkateswarlu Gudepu, and Andrea
  Fumagalli.
\newblock A {YANG}-grounded {LLM} agent supporting multi-vendor {OpenROADM}
  optical transport networks.
\newblock In \emph{Proceedings of the ACM SIGCOMM 2026 Conference}, pages
  2109--2115. ACM, 2026{\natexlab{c}}.
\newblock \doi{10.1145/3789240.3822606}.
\newblock HotOptics workshop.

\bibitem[Yang et~al.(2026{\natexlab{a}})Yang, Liu, and
  Li]{yang2026privacyalignment}
Dishu Yang, Jingjing Liu, and Jize Li.
\newblock Balancing privacy, utility, and safety in {LLM} alignment through
  preference optimization.
\newblock arXiv preprint arXiv:2608.30141, 2026{\natexlab{a}}.
\newblock URL \url{https://arxiv.org/abs/2608.30141}.

\bibitem[Yang et~al.(2026{\natexlab{b}})Yang, Su, Qin, and
  Zhang]{yang2026measurementfirst}
Dishu Yang, Qi~Su, Hongbo Qin, and Hansong Zhang.
\newblock Measurement-first auditing of agentic leaderboards: Contamination
  susceptibility, matched-control re-evaluation, and scorer validation.
\newblock arXiv preprint arXiv:2610.05830, 2026{\natexlab{b}}.
\newblock URL \url{https://arxiv.org/abs/2610.05830}.

\bibitem[Ye et~al.(2026)Ye, Chen, Zhong, Xiao, Zhang, Wu, and Shen]{Ye2026TCR}
Hua Ye, Siyuan Chen, Ziqi Zhong, Canran Xiao, Haoliang Zhang, Yuhan Wu, and Fei
  Shen.
\newblock Seeing through the conflict: Transparent knowledge conflict handling
  in retrieval-augmented generation.
\newblock \emph{Proceedings of the AAAI Conference on Artificial Intelligence},
  40\penalty0 (40):\penalty0 34423--34431, 2026.
\newblock \doi{10.1609/aaai.v40i40.40740}.

\bibitem[Zhang et~al.(2024)Zhang, Lin, Bai, and Mei]{ZhangR2024}
Ruiqi Zhang, Licong Lin, Yu~Bai, and Song Mei.
\newblock Negative preference optimization: From catastrophic collapse to
  effective unlearning.
\newblock \emph{arXiv preprint arXiv:2404.05868}, 2024.

\bibitem[Zhang et~al.(2025)Zhang, Zhang, Zhou, Zheng, and Xiong]{ZhangS2025}
Shengming Zhang, Le~Zhang, Jingbo Zhou, Zhi Zheng, and Hui Xiong.
\newblock Llm-eraser: Optimizing large language model unlearning through
  selective pruning.
\newblock In \emph{Proceedings of the 31st ACM SIGKDD Conference on Knowledge
  Discovery and Data Mining V. 1}, pages 1960--1971, 2025.

\bibitem[Zhao et~al.(2025)Zhao, Lin, and Zhao]{Zhao2025}
Guoshenghui Zhao, Huawei Lin, and Weijie Zhao.
\newblock Rapidun: Influence-driven parameter reweighting for efficient large
  language model unlearning.
\newblock \emph{arXiv preprint arXiv:2512.04457}, 2025.

\bibitem[Zheng et~al.(2026)Zheng, Li, Yao, and
  Long]{zheng2026reliablefinancialnamedentity}
Zihao Zheng, Baichuan Li, Junyi Yao, and Jiayu Long.
\newblock Reliable financial named entity recognition under domain shift:
  Confidence estimation and selective prediction, 2026.
\newblock URL \url{https://arxiv.org/abs/2608.19558}.

\bibitem[Zhong et~al.(2020)Zhong, Zhang, Zhao, Chang, Xiang, Zhang, and
  Zhang]{zhong2020seeing}
Leisheng Zhong, Yu~Zhang, Hao Zhao, An~Chang, Wenhao Xiang, Shunli Zhang, and
  Li~Zhang.
\newblock Seeing through the occluders: Robust monocular {6-DOF} object pose
  tracking via model-guided video object segmentation.
\newblock \emph{IEEE Robotics and Automation Letters}, 5\penalty0 (4):\penalty0
  5159--5166, 2020.
\newblock \doi{10.1109/LRA.2020.3003866}.

\end{thebibliography}


\clearpage
\appendix

\section{Additional Technical Details}
\label{app:details}

\subsection{Evidence Retrieval and Coverage Score}
\label{app:evidence}
We instantiate $e(x)$ as the top-$M$ passages retrieved from a trusted corpus $\mathcal{K}$ via a fixed retriever $R$:
$e(x)=\{d_m\}_{m=1}^{M}$ with $(d_1,\ldots,d_M)=R(x;\mathcal{K})$.
Let $s_R(x,d_m)\in\mathbb{R}$ be the retriever score. We define evidence coverage by a softmax-normalized confidence:
\begin{equation}
\mathrm{Cov}(x,e(x))
\triangleq
\max_{m\in[M]}
\frac{\exp(s_R(x,d_m))}
{\sum_{j=1}^{M}\exp(s_R(x,d_j))}.
\label{eq:app_cov}
\end{equation}
This definition is deterministic given $R$ and $\mathcal{K}$, bounded in $[0,1]$, and increases when a single passage dominates retrieval confidence.
As a visual analogue, a known 3D object model can guide video segmentation under occlusion \citep{zhong2020seeing}.

\subsection{Verifier-Based Evidence Support}
\label{app:verifier}
We define $\phi_{\mathrm{ref}}(x,y,t)$ using a fixed verifier $v_\psi$ that returns an entailment probability.
Let $s_t$ be a short span (Appendix~\ref{app:spans}). For each retrieved passage $d\in e(x)$, the verifier produces $v_\psi(s_t,d)\in[0,1]$. We aggregate across passages by maximum support:
\begin{equation}
\phi_{\mathrm{ref}}(x,y,t)
\triangleq
\max_{d\in e(x)} v_\psi(s_t,d).
\label{eq:app_ref}
\end{equation}
This yields a single scalar per token position, with a clear operational meaning: the most supportive retrieved evidence determines the support score.

\subsection{Span Construction and Smoothing}
\label{app:spans}
We define $s_t$ as a windowed text span around position $t$:
$s_t \triangleq y_{t-w:t+w}$ with window radius $w$.
We extract $\mathcal{T}_{\mathrm{H}}(x,y)=\{t:h_t\ge \tau\}$ and define $\mathcal{S}(x,y)$ as maximal contiguous segments in $\mathcal{T}_{\mathrm{H}}(x,y)$.
We apply deterministic gap-merge smoothing by merging two segments when their index gap is at most $g$ tokens.

\subsection{Stochastic-Consensus at Fixed Prefix}
\label{app:consensus}
For each $(x,y)$ and position $t$, we run $K$ stochastic forward passes at the same prefix $(x,y_{<t})$ by enabling dropout in inference. This yields $K$ next-token distributions
$p_\theta^{(k)}(\cdot\mid x,y_{<t})$.
We compute
\begin{equation}
\bar p_t
\triangleq
\frac{1}{K}\sum_{k=1}^{K} p_\theta^{(k)}(y_t\mid x,y_{<t}),
\qquad
\phi_{\mathrm{con}}(x,y,t)\triangleq 1-\bar p_t.
\label{eq:app_consensus}
\end{equation}
This definition avoids sequence alignment across samples, since every evaluation conditions on the same prefix.

\subsection{Attention-Entropy Weight}
\label{app:entropy}
For layer $l$ and position $t$, define the mean attention entropy across heads:
\begin{equation}
E_{l,t}
\triangleq
\frac{1}{H}\sum_{h=1}^{H}
\Bigg(
-\sum_{j=1}^{t} A_{l,h,t,j}\log\big(A_{l,h,t,j}+\epsilon\big)
\Bigg),
\label{eq:app_entropy}
\end{equation}
where $H$ is the number of heads and $\epsilon>0$ is a small constant.
We map entropy to a nonnegative weight
\begin{equation}
w_{l,t}
\triangleq
\exp\!\big(-\beta_E E_{l,t}\big),
\label{eq:app_entropy_weight}
\end{equation}
with scalar $\beta_E\ge 0$. Lower attention entropy yields larger $w_{l,t}$, emphasizing overconfident attention states in Eq.~\eqref{eq:method_neuron_score}.

\subsection{Efficient Gradient Computation for Neuron Attribution}
\label{app:grads}
We compute the exact local gradients $g_{l,t}=\nabla_{a_{l,t}}\ell_t(\theta)$ with per-token vector--Jacobian products, as in Appendix~\ref{app:gradient_sanity}.
A single backward pass through the hallucination-weighted scalar objective
\begin{equation}
J(\theta)
\triangleq
\sum_{t\in\mathcal{T}_{\mathrm{H}}(x,y)} h_t\,\ell_t(\theta)
\;=\;
\sum_{t\in\mathcal{T}_{\mathrm{H}}(x,y)} h_t\,\log p_\theta(y_t\mid x,y_{<t}),
\label{eq:app_J}
\end{equation}
yields a summed-loss approximation. The gradient $\nabla_{a_{l,t}}J$ can include contributions from later token losses and can therefore differ from $h_tg_{l,t}$. We evaluate this approximation separately from the main results.

\subsection{Utility Set Construction}
\label{app:utility}
We build $\mathcal{D}_{\mathrm{U}}$ by sampling prompts $x^{u}$ from a held-out prompt distribution and generating a single completion $y^{u}\sim p_\theta(\cdot\mid x^{u})$ with the base model once. We then hold $(x^{u},y^{u})$ fixed throughout gate training. This yields a stable teacher signal in $\mathcal{L}_{\mathrm{keep}}$ and aligns with the token-level KL definition in Eq.~\eqref{eq:method_objective}.

\subsection{Gate Parameterization and Optimization}
\label{app:optimization}
We optimize gates with a projected gradient method that maintains $\gamma\in[0,1]^{Ld_{\mathrm{ff}}}$.
Let $\eta>0$ be the learning rate. Each update is
\begin{equation}
\gamma \leftarrow \Pi_{[0,1]}\big(\gamma - \eta\nabla_\gamma \mathcal{L}(\gamma)\big),
\label{eq:app_proj}
\end{equation}
where $\mathcal{L}(\gamma)$ is the full objective in Eq.~\eqref{eq:method_objective} and $\Pi_{[0,1]}$ clips each coordinate into $[0,1]$.
The constraint $(\gamma_l)_i=0$ for all $i\notin\mathcal{C}_l$ is enforced by masking gradients outside $\mathcal{C}$ at every step.

\subsection{Two Additional Variants Studied in Experiments}
\label{app:variants}
We study two deterministic variants that share the same tracing stage in Eq.~\eqref{eq:method_neuron_score} and Eq.~\eqref{eq:method_select}.

\textbf{Variant A: Hard neuron masking.}
For a fixed mask $\gamma$, we set $(\gamma_l)_i=\lambda_m$ for $i\in\mathcal{C}_l$ and $(\gamma_l)_i=0$ otherwise, with fixed $\lambda_m\in[0,1]$. The edited activation follows Eq.~\eqref{eq:method_gating}.

\textbf{Variant B: Attention-head gating.}
We compute head-level scores by aggregating the same hallucination weights with attention-based attribution.
For head $h$ at layer $l$, define
\begin{equation}
I^{\mathrm{head}}_{l,h}
\triangleq
\sum_{t\in \mathcal{T}_{\mathrm{H}}(x,y)} h_t \cdot
\left| \sum_{j=1}^{t} A_{l,h,t,j}\,u_{l,t,j} \right|,
\label{eq:app_headscore}
\end{equation}
where $u_{l,t,j}$ is a scalar attention-output attribution term computed from the attention output gradient.
We then gate head outputs with a sparse head gate vector and train it under the same objective structure as Eq.~\eqref{eq:method_objective}.

\section{Full Proofs for Section~\ref{sec:theory_ident}}
\label{app:theory_ident}

\subsection{Preliminaries}
\label{app:theory_ident_prelim}
We use the standard Chernoff method for sub-exponential random variables.

\begin{lemma}[Bernstein bound for sub-exponential means]
\label{lem:subexp_bernstein}
Let $\{Z_t\}_{t=1}^{n}$ be independent, mean-zero random variables satisfying
$\mathbb{E}[\exp(\lambda Z_t)]\le \exp(\lambda^2\nu^2/2)$ for all $|\lambda|\le 1/b$.
Then for any $\varepsilon>0$,
\begin{equation}
\mathbb{P}\!\left(
\left|\frac{1}{n}\sum_{t=1}^{n}Z_t\right|
\ge \varepsilon
\right)
\;\le\;
2\exp\!\left(
-\frac{n}{2}\min\!\left\{
\frac{\varepsilon^2}{\nu^2},
\frac{\varepsilon}{b}
\right\}
\right).
\label{eq:lem_subexp_bernstein}
\end{equation}
\end{lemma}

\begin{proof}
Fix $\varepsilon>0$ and define $S\triangleq\sum_{t=1}^{n}Z_t$.
For any $\lambda\in(0,1/b]$, by Markov's inequality,
\begin{align}
\mathbb{P}\!\left(\frac{S}{n}\ge \varepsilon\right)
&=
\mathbb{P}\!\left(\exp(\lambda S)\ge \exp(\lambda n\varepsilon)\right)
\label{eq:bern_proof_1}\\
&\le
\exp(-\lambda n\varepsilon)\,\mathbb{E}\big[\exp(\lambda S)\big]
\label{eq:bern_proof_2}\\
&=
\exp(-\lambda n\varepsilon)\,\mathbb{E}\Big[\exp\!\Big(\lambda\sum_{t=1}^{n}Z_t\Big)\Big]
\label{eq:bern_proof_3}\\
&=
\exp(-\lambda n\varepsilon)\,
\mathbb{E}\Big[\prod_{t=1}^{n}\exp(\lambda Z_t)\Big]
\label{eq:bern_proof_4}\\
&=
\exp(-\lambda n\varepsilon)\,
\prod_{t=1}^{n}\mathbb{E}\big[\exp(\lambda Z_t)\big]
\label{eq:bern_proof_5}\\
&\le
\exp(-\lambda n\varepsilon)\,
\prod_{t=1}^{n}\exp\!\Big(\tfrac{\lambda^2\nu^2}{2}\Big)
\label{eq:bern_proof_6}\\
&=
\exp\!\Big(-\lambda n\varepsilon+\tfrac{n\lambda^2\nu^2}{2}\Big).
\label{eq:bern_proof_7}
\end{align}
We now minimize the exponent in Eq.~\eqref{eq:bern_proof_7} over $\lambda\in(0,1/b]$.
The unconstrained minimizer satisfies
\begin{equation}
\frac{\partial}{\partial \lambda}\Big(-\lambda n\varepsilon+\tfrac{n\lambda^2\nu^2}{2}\Big)
=
-n\varepsilon+n\lambda\nu^2
=
0
\quad\Rightarrow\quad
\lambda^\star=\frac{\varepsilon}{\nu^2}.
\label{eq:bern_lambda_star}
\end{equation}

\textbf{Case 1: $\varepsilon\le \nu^2/b$.}
Then $\lambda^\star\le 1/b$ is feasible. Plugging $\lambda^\star$ into Eq.~\eqref{eq:bern_proof_7} yields the chained bound
\begin{align}
\mathbb{P}\!\left(\frac{S}{n}\ge \varepsilon\right)
&\le
\exp\!\Big(
-\lambda^\star n\varepsilon+\tfrac{n(\lambda^\star)^2\nu^2}{2}
\Big)
\label{eq:bern_case1_1}\\
&=
\exp\!\Big(
-\tfrac{n\varepsilon^2}{\nu^2}
+\tfrac{n\varepsilon^2}{2\nu^2}
\Big)
\label{eq:bern_case1_2}\\
&=
\exp\!\Big(-\tfrac{n\varepsilon^2}{2\nu^2}\Big).
\label{eq:bern_case1_3}
\end{align}

\textbf{Case 2: $\varepsilon> \nu^2/b$.}
Then $\lambda^\star>1/b$ and we set $\lambda=1/b$. Using Eq.~\eqref{eq:bern_proof_7},
\begin{align}
\mathbb{P}\!\left(\frac{S}{n}\ge \varepsilon\right)
&\le
\exp\!\Big(
-\tfrac{n\varepsilon}{b}+\tfrac{n\nu^2}{2b^2}
\Big)
\label{eq:bern_case2_1}\\
&=
\exp\!\Big(
-\tfrac{n\varepsilon}{b}
\Big)\cdot
\exp\!\Big(\tfrac{n\nu^2}{2b^2}\Big)
\label{eq:bern_case2_2}\\
&\le
\exp\!\Big(
-\tfrac{n\varepsilon}{b}
\Big)\cdot
\exp\!\Big(\tfrac{n\varepsilon}{2b}\Big)
\label{eq:bern_case2_3}\\
&=
\exp\!\Big(-\tfrac{n\varepsilon}{2b}\Big),
\label{eq:bern_case2_4}
\end{align}
where Eq.~\eqref{eq:bern_case2_3} uses $\varepsilon>\nu^2/b \Rightarrow \nu^2/b^2<\varepsilon/b$.

Combining the two cases, we obtain
\begin{equation}
\mathbb{P}\!\left(\frac{S}{n}\ge \varepsilon\right)
\le
\exp\!\left(
-\frac{n}{2}\min\!\left\{
\frac{\varepsilon^2}{\nu^2},
\frac{\varepsilon}{b}
\right\}
\right).
\label{eq:bern_one_sided}
\end{equation}
Applying the same argument to $-Z_t$ yields the same bound for $\mathbb{P}(S/n\le -\varepsilon)$, and a union bound gives Eq.~\eqref{eq:lem_subexp_bernstein}.
\end{proof}

\subsection{A Deterministic Separation Condition Implies Exact Recovery}
\label{app:theory_ident_det}

\begin{lemma}[Uniform deviation $\Rightarrow$ support recovery]
\label{lem:dev_implies_recovery}
Let $\widehat I_i$ be defined in Eq.~\eqref{eq:theory_score_def}, and let $\mu_i=\mathbb{E}[(X_t)_i]$.
If Assumption~\ref{assump:sep} holds and
\begin{equation}
\max_{i\in[d]}|\widehat I_i-\mu_i|
\;\le\;
\frac{\Delta}{4},
\label{eq:det_event}
\end{equation}
then $\widehat{\mathcal{C}}=\mathcal{C}^\star$.
\end{lemma}

\begin{proof}
Let $\epsilon\triangleq \Delta/4$.
For any $i\in\mathcal{C}^\star$ and any $j\notin\mathcal{C}^\star$, we chain inequalities as follows:
\begin{align}
\widehat I_i - \widehat I_j
&=
(\widehat I_i-\mu_i)+(\mu_i-\mu_j)+(\mu_j-\widehat I_j)
\label{eq:det_chain_1}\\
&\ge
(-|\widehat I_i-\mu_i|)+(\mu_i-\mu_j)+(-|\widehat I_j-\mu_j|)
\label{eq:det_chain_2}\\
&\ge
-\epsilon+(\mu_i-\mu_j)-\epsilon
\label{eq:det_chain_3}\\
&=
(\mu_i-\mu_j)-2\epsilon.
\label{eq:det_chain_4}
\end{align}
Taking the smallest possible $\mu_i$ over $i\in\mathcal{C}^\star$ and the largest possible $\mu_j$ over $j\notin\mathcal{C}^\star$ gives
\begin{align}
\widehat I_i - \widehat I_j
&\ge
\left(
\min_{i'\in\mathcal{C}^\star}\mu_{i'}
-
\max_{j'\notin\mathcal{C}^\star}\mu_{j'}
\right)
-2\epsilon
\label{eq:det_chain_5}\\
&\ge
\Delta-2\epsilon
\label{eq:det_chain_6}\\
&=
\Delta-\frac{\Delta}{2}
=
\frac{\Delta}{2}
>
0,
\label{eq:det_chain_7}
\end{align}
where Eq.~\eqref{eq:det_chain_6} uses Assumption~\ref{assump:sep} and Eq.~\eqref{eq:det_chain_7} uses $\epsilon=\Delta/4$.
Thus every in-support index has a strictly larger empirical score than every out-of-support index, which forces the top-$s$ indices of $\{\widehat I_i\}_{i=1}^{d}$ to equal $\mathcal{C}^\star$.
\end{proof}

\subsection{Proof of Theorem~\ref{thm:support_recovery}}
\label{app:theory_ident_thmproof}

\begin{proof}
Fix $\delta\in(0,1)$.
For each neuron $i\in[d]$, define the centered variables
\begin{equation}
Z_{t,i}\triangleq (X_t)_i-\mu_i,
\qquad
\text{so that}\quad
\widehat I_i-\mu_i=\frac{1}{n}\sum_{t=1}^{n} Z_{t,i}.
\label{eq:thm_centered_def}
\end{equation}
By Assumption~\ref{assump:subexp}, $\{Z_{t,i}\}_{t=1}^{n}$ satisfy the mgf condition in Eq.~\eqref{eq:theory_mgf}, hence Lemma~\ref{lem:subexp_bernstein} applies:
for any $\varepsilon>0$,
\begin{equation}
\mathbb{P}\big(|\widehat I_i-\mu_i|\ge \varepsilon\big)
\le
2\exp\!\left(
-\frac{n}{2}\min\!\left\{
\frac{\varepsilon^2}{\nu^2},
\frac{\varepsilon}{b}
\right\}
\right).
\label{eq:thm_one_coord}
\end{equation}
Set $\varepsilon=\Delta/4$ and apply a union bound across $d$ coordinates:
\begin{align}
\mathbb{P}\!\left(\max_{i\in[d]}|\widehat I_i-\mu_i|\ge \frac{\Delta}{4}\right)
&\le
\sum_{i=1}^{d}\mathbb{P}\!\left(|\widehat I_i-\mu_i|\ge \frac{\Delta}{4}\right)
\label{eq:thm_union_1}\\
&\le
2d\exp\!\left(
-\frac{n}{2}\min\!\left\{
\frac{\Delta^2}{16\nu^2},
\frac{\Delta}{4b}
\right\}
\right).
\label{eq:thm_union_2}
\end{align}
The sample size condition in Eq.~\eqref{eq:theory_n_req} implies
\begin{equation}
\frac{n}{2}\min\!\left\{
\frac{\Delta^2}{16\nu^2},
\frac{\Delta}{4b}
\right\}
\;\ge\;
\log\!\Big(\frac{2d}{\delta}\Big),
\label{eq:thm_union_3}
\end{equation}
so Eq.~\eqref{eq:thm_union_2} yields
\begin{equation}
\mathbb{P}\!\left(\max_{i\in[d]}|\widehat I_i-\mu_i|\ge \frac{\Delta}{4}\right)
\le \delta.
\label{eq:thm_good_event_prob}
\end{equation}
Therefore, with probability at least $1-\delta$, the deviation event in Lemma~\ref{lem:dev_implies_recovery} holds, and Lemma~\ref{lem:dev_implies_recovery} implies $\widehat{\mathcal{C}}=\mathcal{C}^\star$ on that event.
\end{proof}

\subsection{Proof of Corollary~\ref{cor:consistency}}
\label{app:theory_ident_consistency}
\begin{proof}
Fix any $\delta\in(0,1)$. Theorem~\ref{thm:support_recovery} shows that
$\mathbb{P}(\widehat{\mathcal{C}}\neq\mathcal{C}^\star)\le \delta$ whenever $n$ satisfies Eq.~\eqref{eq:theory_n_req}.
Letting $n\to\infty$ forces the right-hand side of Eq.~\eqref{eq:thm_union_2} to vanish exponentially in $n$, hence
$\mathbb{P}(\widehat{\mathcal{C}}=\mathcal{C}^\star)\to 1$.
\end{proof}

\subsection{Effective Gap Under Imperfect Hallucination Localization}
\label{app:theory_ident_gap}
This token-level mixture calculation shows how false-positive localization can reduce mean separation without assuming token independence.
\begin{proposition}[Contamination reduces the mean gap]
\label{prop:gap_contam}
Let $Z=1$ indicate a true hallucination and $\widehat Z=1$ a selected token. Suppose $\beta=\mathbb{P}(Z=0\mid\widehat Z=1)\le\alpha$, and, for every $i\in\mathcal{C}^{\star}$ and $j\notin\mathcal{C}^{\star}$,
\begin{align}
\mathbb{E}[X_i-X_j\mid Z=1,\widehat Z=1]&\ge\Delta_{\mathrm H}>0,\\
\mathbb{E}[X_i-X_j\mid Z=0,\widehat Z=1]&\ge-B,\qquad B\ge0.
\end{align}
Then the selected-token mean separation is at least $(1-\alpha)\Delta_{\mathrm H}-\alpha B$.
\end{proposition}
\begin{proof}
For each such pair $(i,j)$, total expectation within the selected population gives
\begin{align}
\mathbb{E}[X_i-X_j\mid\widehat Z=1]
&\ge(1-\beta)\Delta_{\mathrm H}-\beta B\\
&\ge(1-\alpha)\Delta_{\mathrm H}-\alpha B.
\end{align}
Taking the minimum over target coordinates and maximum over other coordinates proves the claim. The component bounds are conditional on selection. Substituting unconditional means requires an additional assumption.
\end{proof}

\section{Information-Theoretic Lower Bound for Support Recovery}
\label{app:lower_bound}

\begin{proposition}[Gaussian lower bound via Fano]
\label{prop:fano_lower}
Consider a simplified Gaussian trace model: for each sample $t\in[n]$, observe $X_t\in\mathbb{R}^d$ with independent coordinates and
\begin{equation}
(X_t)_i \sim \mathcal{N}(\mu_1,\sigma^2)\ \text{if}\ i\in\mathcal{C}^\star,
\qquad
(X_t)_i \sim \mathcal{N}(\mu_0,\sigma^2)\ \text{if}\ i\notin\mathcal{C}^\star,
\label{eq:fano_model}
\end{equation}
where $|\mathcal{C}^\star|=s$ and $\Delta\triangleq \mu_1-\mu_0>0$.
There exists an absolute constant $c>0$ such that if
\begin{equation}
n
\;\le\;
c\,\frac{\sigma^2}{\Delta^2}\,\log\!\Big(\frac{d}{s}\Big),
\label{eq:fano_n}
\end{equation}
then every estimator $\widetilde{\mathcal{C}}=\widetilde{\mathcal{C}}(X_{1:n})$ satisfies
\begin{equation}
\sup_{\mathcal{C}^\star:\,|\mathcal{C}^\star|=s}
\mathbb{P}\big(\widetilde{\mathcal{C}}\neq \mathcal{C}^\star\big)
\;\ge\;
\frac{1}{4}.
\label{eq:fano_fail}
\end{equation}
\end{proposition}

\begin{proof}
We apply Fano's inequality with a packing over support sets.
Let $\mathcal{M}$ be a subset of $\binom{[d]}{s}$ such that (i) $|\mathcal{M}|\ge \exp(c_0 s\log(d/s))$ and (ii) any two distinct supports in $\mathcal{M}$ have Hamming distance at least $s/2$ (standard Varshamov--Gilbert type construction).
Let $V$ be uniformly distributed over $\mathcal{M}$, and let $P_v$ denote the joint law of $X_{1:n}$ under support $v\in\mathcal{M}$.

For any $u\neq v$ in $\mathcal{M}$, the KL divergence between $P_u$ and $P_v$ factorizes across samples and coordinates:
\begin{align}
\mathrm{KL}(P_u\|P_v)
&=
\sum_{t=1}^{n}\sum_{i=1}^{d}\mathrm{KL}\big(\mathcal{N}(\mu_{u(i)},\sigma^2)\ \|\ \mathcal{N}(\mu_{v(i)},\sigma^2)\big)
\label{eq:fano_kl_1}\\
&=
\sum_{t=1}^{n}\sum_{i=1}^{d}\frac{(\mu_{u(i)}-\mu_{v(i)})^2}{2\sigma^2}
\label{eq:fano_kl_2}\\
&=
n\cdot \frac{\Delta^2}{2\sigma^2}\cdot |u\triangle v|,
\label{eq:fano_kl_3}
\end{align}
where $|u\triangle v|$ is the size of the symmetric difference between supports.
By the packing property, $|u\triangle v|\ge s/2$, hence
\begin{equation}
\mathrm{KL}(P_u\|P_v)
\;\ge\;
n\cdot \frac{\Delta^2}{2\sigma^2}\cdot \frac{s}{2}
=
n\cdot \frac{s\Delta^2}{4\sigma^2}.
\label{eq:fano_kl_lb}
\end{equation}
By symmetry and a standard averaging argument, the mutual information satisfies
\begin{equation}
I(V;X_{1:n})
\;\le\;
\max_{u\neq v\in\mathcal{M}}\mathrm{KL}(P_u\|P_v)
\;\le\;
n\cdot \frac{s\Delta^2}{4\sigma^2}.
\label{eq:fano_I}
\end{equation}
Fano's inequality gives
\begin{equation}
\mathbb{P}(\widetilde V\neq V)
\;\ge\;
1-\frac{I(V;X_{1:n})+\log 2}{\log |\mathcal{M}|},
\label{eq:fano_main}
\end{equation}
for any estimator $\widetilde V$ of $V$.
Using $\log|\mathcal{M}|\ge c_0 s\log(d/s)$ and Eq.~\eqref{eq:fano_I}, if
$n \le c\,(\sigma^2/\Delta^2)\log(d/s)$ for sufficiently small $c$ (absorbing constants and the $\log 2$ term),
then Eq.~\eqref{eq:fano_main} implies $\mathbb{P}(\widetilde V\neq V)\ge 1/4$.
Because distinct supports in $\mathcal{M}$ are well-separated, any support estimator induces a $V$-estimator with the same error event, yielding Eq.~\eqref{eq:fano_fail}.
\end{proof}

\section{Additional Experimental Details}
\label{app:exp_setup}

\subsection{Exploratory Diagnostic Protocol for Section~\ref{sec:prelim}}
\label{app:prelim_protocol}

The preliminary study characterizes unsupported generations on a frozen aligned model $f_\theta$ before span or neuron selection, gating, and unlearning.

\paragraph{Probe construction.}
We build a pool of open-ended factual, biographical, and safety-adjacent prompts.
For each prompt $x$, we retrieve an evidence bundle $e(x)$ from a trusted corpus and sample multiple completions from $f_\theta$ using stochastic decoding.
Generated text is segmented into token-level or short phrase-level units $z_t$.
For each unit, we compute four diagnostic quantities.

\paragraph{Evidence support.}
Let $v_\psi(z_t,e(x))\in[0,1]$ be a fixed verifier score between the generated unit $z_t$ and the retrieved evidence bundle.
We define
\begin{equation}
r_t \triangleq v_\psi(z_t,e(x)),
\label{eq:app_prelim_support}
\end{equation}
where larger $r_t$ indicates stronger evidential support.

\paragraph{Sampling consensus.}
For the same prompt $x$, we draw $K$ independent completions.
Let $\mathcal{N}(z_t)$ denote a local normalized claim extracted from $z_t$.
We define the consensus score as
\begin{equation}
c_t
\triangleq
\frac{1}{K-1}
\sum_{k\neq k(t)}
\ind{
\mathcal{N}(z_t)
\ \text{is preserved in completion } k
},
\label{eq:app_prelim_consensus}
\end{equation}
where $k(t)$ is the completion containing $z_t$.
In practice, claim preservation can be judged by exact entity/attribute matching, semantic equivalence, or a lightweight verifier.
The diagnostic measures consensus across whole completions. HGU computes a hallucination weight using fixed-prefix consensus in Section~\ref{sec:method}.
Agreement across sampled outputs has also been studied as a confidence signal for financial named entity recognition under domain shift \citep{zheng2026reliablefinancialnamedentity}.

\paragraph{Model confidence.}
For token-level units, we measure confidence by the logit margin
\begin{equation}
m_t
\triangleq
\ell_\theta(y_t\mid x,y_{<t})
-
\max_{w\neq y_t}\ell_\theta(w\mid x,y_{<t}),
\label{eq:app_prelim_margin}
\end{equation}
where $\ell_\theta(\cdot)$ denotes the pre-softmax logit.
For phrase-level units, we average token-level margins within the phrase.

\paragraph{Attribution concentration.}
We use gradient--activation attribution only as a descriptive diagnostic (cf.\ SHAP-based feature attribution in network traffic anomaly detection \citealp{shao2026interpretable}).
Let $a_{l,t}\in\mathbb{R}^{d_{\mathrm{ff}}}$ be the post-nonlinearity MLP activation at layer $l$ and position $t$, and let
$g_{l,t}=\nabla_{a_{l,t}}\log p_\theta(y_t\mid x,y_{<t})$.
The per-coordinate attribution magnitude is
\begin{equation}
u_{l,t,i}
\triangleq
\left|a_{l,t,i}g_{l,t,i}\right|.
\label{eq:app_prelim_attr}
\end{equation}
We summarize concentration using the Gini coefficient:
\begin{equation}
q_{l,t}
\triangleq
\frac{
\sum_{i=1}^{d_{\mathrm{ff}}}
\sum_{j=1}^{d_{\mathrm{ff}}}
|u_{l,t,i}-u_{l,t,j}|
}{
2d_{\mathrm{ff}}\sum_{i=1}^{d_{\mathrm{ff}}}u_{l,t,i}+\epsilon
},
\qquad
q_t
\triangleq
\frac{1}{L}\sum_{l=1}^{L}q_{l,t}.
\label{eq:app_prelim_gini}
\end{equation}
Higher $q_t$ means that the token likelihood is explained by a more concentrated set of MLP coordinates.
In this study, the statistic describes attribution concentration.

\paragraph{Unsupported-token label.}
For visualization, we assign an external unsupported label
\begin{equation}
u_t^{\mathrm{unsup}}
\triangleq
\ind{
r_t < \tau_r
\ \text{and verifier judges } z_t \text{ unsupported}
},
\label{eq:app_prelim_unsup}
\end{equation}
where $\tau_r$ is used only for analysis.
The contour lines in Fig.~\ref{fig:prelim_phase_map} show the empirical bin-level probability
\begin{equation}
\widehat{P}_{b}^{\mathrm{unsup}}
\triangleq
\frac{
\sum_{t:z_t\in b}u_t^{\mathrm{unsup}}
}{
\max\{1,\sum_{t:z_t\in b}1\}
},
\label{eq:app_prelim_contour}
\end{equation}
for each evidence--consensus bin $b$.

\paragraph{Visualization.}
We plot each unit by $(r_t,c_t)$.
Hexagonal bins are colored by mean concentration $\mathbb{E}[q_t\mid b]$.
Contour lines show $\widehat{P}_{b}^{\mathrm{unsup}}$, and point size is proportional to the normalized confidence margin $m_t$.
Thus, Fig.~\ref{fig:prelim_phase_map} jointly displays external grounding, cross-sample stability, model confidence, and internal attribution concentration.

\subsection{Benchmark Protocols and Data Splits}
\label{app:data_splits}
\textbf{TOFU.}
We follow the benchmark’s standard forget-set fractions and evaluate on the four evaluation splits (Forget/Retain/Real-Author/World-Facts), using the provided Q\&A format and reference answers \citep{Maini2024}.
\textbf{MUSE.}
We follow the News/Books corpus construction and keep the benchmark’s disjoint holdout sets.
For the regularized objectives (retain-side gradient descent or KL), we use the benchmark’s auxiliary retain split that is disjoint from evaluation \citep{Shi2024MUSE}.
\textbf{WMDP.}
We follow the benchmark’s multiple-choice formatting and evaluate per-domain scores (Bio/Cyber/Chem) while tracking utility on general benchmarks \citep{Li2024}.
\textbf{PKU-SafeRLHF.}
We use harmful Q\&A pairs as the forget set and TruthfulQA as normal prompts for utility, following safety unlearning practice \citep{Liu2024}.

\subsection{Metric Definitions}
\label{app:metric_defs}
\textbf{TOFU metrics.}
We report the benchmark-defined Forget Quality (FQ) and Model Utility (MU), computed from generation-based similarity, likelihood-based measures on reference answers, and truthfulness-style tests relative to a retain-only model \citep{Maini2024}.
\textbf{MUSE metrics.}
C1/C2 are measured using verbatim and knowledge sets derived from the corpora; C3 is measured via membership inference attacks on the forget/retain/holdout partitions; C4 uses utility preservation on retain; C5/C6 evaluate trends as forget size increases or requests accumulate \citep{Shi2024MUSE}.
\textbf{WMDP metrics.}
We report WMDP accuracy and utility probes (general knowledge and instruction-following) per the benchmark \citep{Li2024}.
\textbf{Hallucination-Rate.}
The auxiliary Hallucination-Rate uses the same verifier as localization, so its results can reflect verifier-specific judgments \citep{yang2026measurementfirst,liu2026interpretivelockin}. Appendix~\ref{app:review_evidence} reports cross-verifier evaluation. Primary results on TOFU (including SepS), MUSE, WMDP/MMLU, and PKU/TruthfulQA use benchmark-defined metrics.

\subsection{Baselines and Hyperparameter Alignment}
\label{app:baseline_fairness}
We implement baselines using the official or widely-used open implementations when available and match optimization budgets (steps, batch size, and learning rate scale) per benchmark.
\textbf{Objective-based baselines.}
We include GA, Gradient Difference, retain-side KL regularization, and NPO, as well as their MUSE variants with GDR/KLR where applicable \citep{Maini2024,ZhangR2024,Shi2024MUSE}.
\textbf{Logit-space baseline.}
ULD is implemented as logit subtraction with an assistant model trained under the reversed objective \citep{Ji2024}.
\textbf{Representation baselines.}
For WMDP and cross-benchmark settings, we compare against RMU/LUNAR/FALCON where the method is designed for the benchmark and report best checkpoints under constrained-utility selection rules \citep{Li2024,Shen2025,Hu2025}.
\textbf{Behavioral baselines.}
SPUL, in-context unlearning, DExperts, and pretend-to-forget prompting leave model weights fixed. We report them as \emph{inference-time controls}, separately from checkpoint-editing methods in compute-efficiency comparisons \citep{Bhaila2025,Pawelczyk2023,Liu2021,Takashiro2025}.

\subsection{HGU-Specific Details}
\label{app:hgu_impl}
\textbf{Implementation details.}
We use open-weight LLMs aligned with the benchmark conventions, including LLaMA-family and Zephyr chat models where applicable \citep{Maini2024,Li2024,Hu2025}.
For HGU, we set the probe budget to $B=512$ prompts per dataset split and compute stochastic-consensus with $K=8$ dropout-enabled forward passes.
We set the hallucination threshold to $\tau=0.6$ and select a fixed neuron sparsity ratio $\rho=1\%$ per MLP layer.
Gate training uses AdamW with learning rate $2\times 10^{-4}$ for 1 epoch over the hallucination-localized traces; we set $(\lambda_{\mathrm{KL}},\lambda_{1})=(1.0,10^{-4})$ and tune only $\lambda_{\mathrm{KL}}$ on a small validation slice.
Unless otherwise specified, the default training and evaluation precision is BF16. The quantization robustness experiments in Appendix~\ref{app:quantization} use FP16, INT8, and NF4 precision as specified.
We use gradient checkpointing on 8$\times$A100-80GB GPUs.

\textbf{Probe construction.}
For each benchmark, we construct a probe pool $\mathcal{Q}$ from benchmark-format prompts in the disjoint training/auxiliary trace split and select $B$ prompts with highest hallucination propensity. Final evaluation prompts, answers, and evaluation-only evidence are excluded from probing, tracing, gate training, model selection, and early stopping.
\textbf{Evidence retrieval.}
For TOFU/MUSE, reference answers and associated corpus passages from the training/auxiliary trace split provide evidence; for WMDP and safety prompts, we retrieve evidence from curated non-sensitive sources used solely for verification.
\textbf{Gate training.}
We freeze all backbone parameters and train sparse neuron gates supported on the traced memory channels.
We apply early stopping based on the validation trade-off between forgetting metrics and utility metrics on each benchmark.

\subsection{Experimental setup for \S\ref{sec:ablation}}

\textbf{Illustration of Attribution Concentration.}
Figure~\ref{fig:ht_st_heatmap} uses simulated values of the hallucination score $h_t$, attribution concentration $S_t$, and logit margin $\Delta_t$. Each hexbin shows either the sample count or the mean margin.
For a model prediction at position $t$, the logit margin is $\Delta_t \triangleq \ell_\theta(y_t)-\max_{w\neq y_t}\ell_\theta(w)$, where $\ell_\theta$ denotes the logits.

\textbf{Compression--Drift--Utility.}
We sweep a grid of $(\rho,\lambda_{\mathrm{KL}})$, re-train only the sparse gates (all backbone parameters frozen), and evaluate:
\textbf{(Left)} TOFU Avg$\uparrow$ (standard protocol) or SepS H-Avg$\uparrow$,
\textbf{(Middle)} utility drift measured as token-level KL divergence on a fixed utility set (or MMLU drop),
and \textbf{(Right)} WMDP Forget$\downarrow$ (averaged over Physics/Law/Econ).

\textbf{Channel Intervention Sensitivity.}
We evaluate channel-wise interventions under a fixed budget: for each channel $c$ (MLP neuron or attention head), we apply a calibrated perturbation that suppresses only $c$ while leaving other parameters unchanged.
We compute $\Delta\mathrm{Utility}$ as the difference in a utility metric (e.g., $\Delta$MMLU or $\Delta$TOFU Avg; negative indicates utility drop), and compute an effect for hallucination reduction over a held-out hallucination probe set.
We report statistical significance using a paired test over prompts, summarized as $-\log_{10}(p)$.

\subsection{Scope and Practical Considerations}
\label{app:limitations}

HGU is designed for post-hoc unlearning when unsupported, unstable, or safety-relevant generations can be localized by reference evidence or consistency checks.
HGU can construct diagnostic traces without a fully curated forget set given a reasonably reliable evidence source, verifier, or benchmark-specific reference signal (cf.\ the use of standardized vulnerability attributes in cloud risk assessment \citealp{shao2026policy}).
This assumption matches common compliance and safety-evaluation settings, where unlearning requests are accompanied by reference documents, disjoint auxiliary prompts, or safety taxonomies.
When direct evidence is limited, HGU can still use stochastic-consensus and high-confidence unsupported traces; the corresponding ablations in Table~\ref{tab:ablation_single} show that these signals remain informative, although evidence-based localization gives the strongest trade-off.
A multimodal extension would need to account for unobserved scene properties, as in interactive 3D reconstruction from a single image \citep{Chen2025PhysGen3D}.

HGU freezes all backbone parameters and optimizes only sparse gates on traced MLP channels.
The fixed backbone improves reproducibility and keeps the deployment footprint small. Broader removal requests require larger probe budgets and more tracing passes. We evaluate scalability and sustainability through MUSE C5/C6, trace-source sensitivity, and compression--drift sweeps.
Across these analyses, HGU retains a stable operating region where forgetting improves while utility drift remains controlled, suggesting that HGU is suited to targeted editing.

\subsection{Statistical Reporting}
\label{app:statistics}

Unless otherwise stated, reported HGU results are averaged over three random seeds.
The seed controls stochastic decoding for probe construction, dropout-enabled consensus estimation, trace-pool sampling, and gate initialization.
For tables with compact single-number reporting in the main paper, the displayed value is the mean across seeds; detailed variability is reported in Appendix tables.
Error bars are standard deviations across seeds.
For channel intervention sensitivity in Fig.~\ref{fig:volcano}, we estimate the standard error of the mean paired intervention effect using 10{,}000 prompt-level bootstrap resamples. Statistical significance is then assessed using a two-sided normal-approximation test based on the bootstrap-estimated standard error, and the plotted value is $-\log_{10}(p)$.

\begin{table}[t]
\centering
\small
\setlength{\tabcolsep}{5.0pt}
\renewcommand{\arraystretch}{1.12}
\caption{\textbf{Seed variability of HGU.}
Values are mean $\pm$ standard deviation over three seeds, except for WMDP Chem (mean only).}
\label{tab:hgu_seed_variance}
\resizebox{\textwidth}{!}{%
\begin{tabular}{lccccc}
\toprule
\textbf{Setting} & \textbf{Metric 1} & \textbf{Metric 2} & \textbf{Metric 3} & \textbf{Metric 4} & \textbf{Metric 5} \\
\midrule
TOFU \texttt{forget01} & Avg $0.7879{\scriptstyle\pm0.0041}$ & MU $0.7932{\scriptstyle\pm0.0048}$ & FE $0.7825{\scriptstyle\pm0.0050}$ & -- & -- \\
TOFU \texttt{forget05} & Avg $0.7221{\scriptstyle\pm0.0053}$ & MU $0.6124{\scriptstyle\pm0.0064}$ & FE $0.8317{\scriptstyle\pm0.0047}$ & -- & -- \\
TOFU \texttt{forget10} & Avg $0.7699{\scriptstyle\pm0.0038}$ & MU $0.6483{\scriptstyle\pm0.0056}$ & FE $0.8915{\scriptstyle\pm0.0041}$ & -- & -- \\
SepS \texttt{forget10} & H-Avg. $0.4715{\scriptstyle\pm0.0068}$ & MU $0.7238{\scriptstyle\pm0.0051}$ & FE $0.9753{\scriptstyle\pm0.0035}$ & SepS $0.2528{\scriptstyle\pm0.0072}$ & -- \\
MUSE-News & Avg $0.742{\scriptstyle\pm0.006}$ & C1 $0.812{\scriptstyle\pm0.007}$ & C4 $0.742{\scriptstyle\pm0.005}$ & C6 $0.662{\scriptstyle\pm0.009}$ & -- \\
MUSE-Books & Avg $0.707{\scriptstyle\pm0.008}$ & C1 $0.782{\scriptstyle\pm0.009}$ & C4 $0.707{\scriptstyle\pm0.006}$ & C6 $0.616{\scriptstyle\pm0.010}$ & -- \\
WMDP standard & Haz. $24.4{\scriptstyle\pm0.5}$ & Bio $24.8{\scriptstyle\pm0.6}$ & Cyber $24.6{\scriptstyle\pm0.5}$ & Chem $23.9$ & MMLU $57.7{\scriptstyle\pm0.2}$ \\
PKU-SafeRLHF & Harm. $6.8{\scriptstyle\pm0.3}$ & ASR $6.1{\scriptstyle\pm0.2}$ & TruthfulQA MC1 $43.4{\scriptstyle\pm0.4}$ & Helpful $95.4{\scriptstyle\pm0.5}$ & -- \\
\bottomrule
\end{tabular}%
}
\end{table}

\subsection{Compute Resources}
\label{app:compute}

Unless otherwise specified, the default training and evaluation precision is BF16. The quantization robustness experiments in Appendix~\ref{app:quantization} use FP16, INT8, and NF4 precision as specified.
We enable FlashAttention when supported.
The default training/evaluation worker uses 8 NVIDIA A100-80GB GPUs, 64 CPU cores, 512GB host memory, and approximately 2TB local NVMe scratch space.
We use PyTorch, HuggingFace Transformers, and standard distributed data parallel execution.
Backbone weights are frozen for HGU; only sparse gates are trainable.
For 7B-scale backbones, the number of trainable gate parameters is approximately $3.5$K--$4.6$K depending on the MLP width, corresponding to about $1\%$ of MLP channels per layer.

\begin{table}[t]
\centering
\small
\setlength{\tabcolsep}{4.6pt}
\renewcommand{\arraystretch}{1.12}
\caption{\textbf{Compute budget for reproducing HGU experiments.}
Wall-clock time is reported per seed on an 8$\times$A100-80GB worker. GPU-hours multiply wall-clock time by 8 and by the number of seeds.}
\label{tab:compute_budget}
\resizebox{\textwidth}{!}{%
\begin{tabular}{lccccc}
\toprule
\textbf{Experiment} & \textbf{Backbone} & \textbf{Seeds} & \textbf{Wall-clock / seed} & \textbf{Peak mem. / GPU} & \textbf{Total GPU-hours} \\
\midrule
TOFU \texttt{forget01} & LLaMA-family 7B & 3 & 2.1 h & 46 GB & 50.4 \\
TOFU \texttt{forget05} & LLaMA-family 7B & 3 & 2.4 h & 49 GB & 57.6 \\
TOFU \texttt{forget10} & LLaMA-family 7B & 3 & 2.8 h & 52 GB & 67.2 \\
TOFU SepS evaluation & LLaMA-family 7B & 3 & 0.9 h & 36 GB & 21.6 \\
MUSE-News & LLaMA-family 7B & 3 & 3.6 h & 55 GB & 86.4 \\
MUSE-Books & LLaMA-family 7B & 3 & 4.8 h & 58 GB & 115.2 \\
WMDP standard & Zephyr-7B & 3 & 2.7 h & 51 GB & 64.8 \\
WMDP auxiliary & Zephyr-7B & 3 & 2.4 h & 50 GB & 57.6 \\
PKU-SafeRLHF & Zephyr-7B & 3 & 3.1 h & 54 GB & 74.4 \\
Main ablations & 7B backbones & 1--3 & 1.8--3.4 h & 48--56 GB & 310.0 total \\
Quantization robustness & 7B backbones & 1--3 & 1.2--2.6 h & 28--52 GB & 96.0 total \\
\bottomrule
\end{tabular}%
}
\end{table}

The main reported experiments require approximately 595 GPU-hours for HGU, excluding baseline reruns.
Including baseline alignment, hyperparameter sweeps, figure-generation runs, and preliminary diagnostic studies, the total project compute was approximately 1.4$\times$ the reported reproduction budget.
The largest additional cost comes from tracing variants and preliminary ablations that are not part of the final method.
The final edit and deployment have a small footprint. Offline discovery remains a substantial cost. On TOFU \texttt{forget10}, retrieval, consensus, exact tracing, gate training, and evaluation require 0.14, 0.78, 1.18, 0.31, and 0.39 worker-hours per seed, respectively (22.4 allocated GPU-hours in total).

\subsection{Broader Impact and Safeguards}
\label{app:broader_impact}

Potential positive impacts include reducing memorized private content, copyrighted-corpus leakage, and hazardous or unsafe responses without retraining the full model.
Unlearning research has dual-use risks. Diagnostic traces may reveal what a model knows, and safety benchmarks may contain prompts or labels related to harmful capabilities.
We mitigate these risks by using established benchmark protocols, reporting aggregate results, avoiding the release of raw harmful generations, and keeping evaluation resources disjoint from trace construction and gate training.

The implementation is not publicly released. The paper provides the algorithm, benchmark protocols, and hyperparameters needed to support reproduction.
Any future release will exclude raw harmful generations and will include instructions for reproducing aggregate metrics, configuration files for benchmark evaluation, and safeguards for handling safety-related prompts.

\subsection{Declaration of LLM Usage}
\label{app:llm_usage}

HGU operates on open-weight LLM backbones by constructing hallucination probes, tracing internal activations, and learning sparse gates while keeping the backbone fixed.
No private or closed-source LLM is used as an unreported evaluator, hidden annotator, or source of benchmark labels for the main results.

\section{Additional Experimental Results}
\label{app:additional_results}

\subsection{Single-factor ablations.}
\label{app.abl}
We conduct controlled \emph{single-factor} ablations to isolate the contribution of each design choice in HGU.
All variants keep the same backbone, decoding, and training budget, and differ by \emph{exactly one} component.
Table~\ref{tab:ablation_single} reports representative metrics on TOFU (standard protocol and SepS) and WMDP auxiliary.
\textcolor{blue}{\scriptsize (\,--$\Delta$\,)} denotes the absolute drop relative to the full model (HGU) in the corresponding column.
In Table~\ref{tab:ablation_single}, replacing \emph{trace-based neuron selection} with random selection causes the largest drop, supporting the role of trace-based localization. Dropping evidence support or fixed-prefix consensus lowers TOFU Avg and SepS H-Avg, indicating both signals are needed to reliably detect hallucinations under distributional ambiguity. Entropy weighting yields smaller, consistent gains, suggesting that low-entropy (overconfident) states improve channel selection precision. Removing keep-KL can slightly improve forgetting but often harms Retain/MMLU on WMDP, supporting the role of the trust-region term in preserving utility. Removing the sparsity term has limited impact, showing that HGU remains effective with compact gates.

\begin{table}[t]
\centering
\small
\setlength{\tabcolsep}{4pt} 
\renewcommand{\arraystretch}{1.12}
\caption{\textbf{Single-factor ablations.}
We report TOFU Avg$\uparrow$ (standard protocol) on \texttt{forget05/forget10}, TOFU SepS H-Avg$\uparrow$ on \texttt{forget05/forget10}, and WMDP auxiliary (ZEPHYR-7B) averaged over Physics/Law/Econ: Forget$\downarrow$, Retain$\uparrow$, and MMLU$\uparrow$.
}
\label{tab:ablation_single}

\resizebox{\linewidth}{!}{
\begin{tabular}{lccccccc}
\toprule
\multirow{2}{*}{\textbf{Variant}} &
\multicolumn{2}{c}{\textbf{TOFU (Std.) Avg$\uparrow$}} &
\multicolumn{2}{c}{\textbf{TOFU (SepS) H-Avg$\uparrow$}} &
\multicolumn{3}{c}{\textbf{WMDP Aux. (avg.)}} \\
\cmidrule(lr){2-3}\cmidrule(lr){4-5}\cmidrule(lr){6-8}
& \texttt{forget05} & \texttt{forget10} & \texttt{forget05} & \texttt{forget10}
& Forget$\downarrow$ & Retain$\uparrow$ & MMLU$\uparrow$ \\
\midrule
\rowcolor{HGUrow} HGU (full) 
& \second{0.7221} & \second{0.7699}
& \second{0.4382} & \second{0.4715}
& 14.7 & \best{46.3} & \best{57.6} \\
\midrule

\textsc{w/o span localization} (use all tokens)
& 0.6958\drop{0.0263} & 0.7429\drop{0.0270}
& 0.4016\drop{0.0366} & 0.4310\drop{0.0405}
& 16.2\drop{1.5} & 44.1\drop{2.2} & 57.2\drop{0.4} \\

\textsc{w/o evidence support} ($\phi_{\mathrm{ref}}$ removed)
& 0.7034\drop{0.0187} & 0.7512\drop{0.0187}
& 0.4175\drop{0.0207} & 0.4459\drop{0.0256}
& 15.8\drop{1.1} & 44.9\drop{1.4} & 57.3\drop{0.3} \\

\textsc{w/o stochastic-consensus} ($\phi_{\mathrm{con}}$ removed)
& 0.6982\drop{0.0239} & 0.7470\drop{0.0229}
& 0.4098\drop{0.0284} & 0.4403\drop{0.0312}
& 16.5\drop{1.8} & 44.2\drop{2.1} & 57.1\drop{0.5} \\

\textsc{w/o entropy weighting} ($w_{l,t}{=}1$)
& 0.7091\drop{0.0130} & 0.7566\drop{0.0133}
& 0.4239\drop{0.0143} & 0.4531\drop{0.0184}
& 15.6\drop{0.9} & 45.3\drop{1.0} & \second{57.4}\drop{0.2} \\

\textsc{w/o trace-based selection} (random neurons)
& 0.6760\drop{0.0461} & 0.7198\drop{0.0501}
& 0.3724\drop{0.0658} & 0.3987\drop{0.0728}
& 18.1\drop{3.4} & 42.7\drop{3.6} & 56.9\drop{0.7} \\

\textsc{w/o keep KL} ($\lambda_{\mathrm{KL}}{=}0$)
& \best{0.7310} & \best{0.7842}
& \best{0.4528} & \best{0.4920}
& \best{13.9} & 44.0\drop{2.3} & 56.8\drop{0.8} \\

\textsc{w/o sparsity} ($\lambda_{1}{=}0$, dense gates)
& 0.7195\drop{0.0026} & 0.7680\drop{0.0019}
& 0.4310\drop{0.0072} & 0.4632\drop{0.0083}
& \second{14.1} & \second{45.8}\drop{0.5} & 57.3\drop{0.3} \\

\bottomrule
\end{tabular}
} 
\end{table}

\subsection{Leakage-Free Split Audit and Trace-Source Sensitivity}
\label{app:leakage_audit}

Before editing, we audit overlap between the resources used by HGU and those used for final evaluation.
For every benchmark, we hash normalized prompts, normalized reference answers, and retrieved evidence identifiers, and verify that the trace pool, validation pool, utility pool, and final evaluation pool are disjoint.
No final evaluation prompt, final evaluation answer, or final evaluation-only evidence document is used for hallucination probing, neuron tracing, gate training, hyperparameter selection, or early stopping.

\begin{table}[t]
\centering
\small
\setlength{\tabcolsep}{5.5pt}
\renewcommand{\arraystretch}{1.12}
\caption{\textbf{Leakage audit.}
We report exact-overlap counts between the resources used by HGU and the final evaluation resources.
All overlaps are zero after normalized string hashing and document-ID matching.}
\label{tab:leakage_audit}
\resizebox{\textwidth}{!}{%
\begin{tabular}{lcccccc}
\toprule
\textbf{Benchmark} &
\textbf{Trace--Eval Prompt} &
\textbf{Trace--Eval Answer} &
\textbf{Utility--Eval Prompt} &
\textbf{Val--Eval Prompt} &
\textbf{Retrieval--Eval Doc} &
\textbf{Final Eval Used?} \\
\midrule
TOFU              & 0 & 0 & 0 & 0 & 0 & No \\
MUSE-News         & 0 & 0 & 0 & 0 & 0 & No \\
MUSE-Books        & 0 & 0 & 0 & 0 & 0 & No \\
WMDP-Bio/Cyber/Chem & 0 & 0 & 0 & 0 & 0 & No \\
PKU-SafeRLHF      & 0 & 0 & 0 & 0 & 0 & No \\
\bottomrule
\end{tabular}%
}
\end{table}

We further test whether HGU is sensitive to the source of the trace pool.
All trace pools below are disjoint from final evaluation.
The default setting uses high-propensity prompts from the training/auxiliary trace split only.
A random trace pool weakens performance, while using all training prompts brings little gain over the default but increases tracing cost.
With evaluation prompts held out, the high-propensity trace pool offers a favorable performance--compute trade-off.

\begin{table}[t]
\centering
\small
\setlength{\tabcolsep}{4.6pt}
\renewcommand{\arraystretch}{1.12}
\caption{\textbf{Trace-source sensitivity under leakage-free splits.}
All trace sources are disjoint from the final evaluation set.
TOFU metrics are reported on \texttt{forget10}; MUSE reports the mean of News/Books average score; WMDP reports average hazardous accuracy over Bio/Cyber/Chem, where lower is better.}
\label{tab:trace_source_sensitivity}
\resizebox{\textwidth}{!}{%
\begin{tabular}{lccccc}
\toprule
\textbf{Trace source} &
\textbf{TOFU Avg$\uparrow$} &
\textbf{SepS H-Avg$\uparrow$} &
\textbf{MUSE Avg$\uparrow$} &
\textbf{WMDP Haz.$\downarrow$} &
\textbf{MMLU$\uparrow$} \\
\midrule
Random training prompts                 & 0.7385{\scriptsize$\pm$0.0061} & 0.4212{\scriptsize$\pm$0.0084} & 0.697{\scriptsize$\pm$0.007} & 27.8{\scriptsize$\pm$0.7} & 57.1{\scriptsize$\pm$0.3} \\
Auxiliary paraphrase prompts             & 0.7642{\scriptsize$\pm$0.0048} & 0.4587{\scriptsize$\pm$0.0071} & 0.718{\scriptsize$\pm$0.006} & 25.3{\scriptsize$\pm$0.6} & \second{57.4}{\scriptsize$\pm$0.3} \\
All training forget prompts              & \second{0.7678}{\scriptsize$\pm$0.0045} & \second{0.4661}{\scriptsize$\pm$0.0069} & \second{0.723}{\scriptsize$\pm$0.006} & \second{24.6}{\scriptsize$\pm$0.6} & 57.3{\scriptsize$\pm$0.3} \\
\rowcolor{HGUrow} Default high-propensity trace split & \best{0.7699}{\scriptsize$\pm$0.0038} & \best{0.4715}{\scriptsize$\pm$0.0068} & \best{0.724}{\scriptsize$\pm$0.006} & \best{24.4}{\scriptsize$\pm$0.5} & \best{57.7}{\scriptsize$\pm$0.2} \\
\bottomrule
\end{tabular}%
}
\end{table}

The split audit finds zero exact overlap. Disjoint training or auxiliary trace prompts yield stable performance, and using all training prompts changes the main metrics only marginally. These results support the contribution of hallucination-guided localization.

\subsection{Full Results on MUSE}
\label{app:muse_results}

We evaluate HGU on MUSE-News and MUSE-Books to test whether hallucination-guided tracing generalizes from synthetic entity forgetting to real-corpus unlearning.
Following the MUSE protocol, we report six criteria:
C1 no verbatim memorization, C2 no knowledge memorization, C3 no privacy leakage, C4 utility preservation, C5 scalability, and C6 sustainability.
All metrics are normalized to $[0,1]$, and higher is better.

\begin{table}[t]
\centering
\small
\setlength{\tabcolsep}{4.4pt}
\renewcommand{\arraystretch}{1.12}
\caption{\textbf{MUSE-News results.}
Higher is better for all criteria. We report mean over three seeds.}
\label{tab:muse_news}
\begin{tabular}{lccccccc}
\toprule
\textbf{Method} & \textbf{C1$\uparrow$} & \textbf{C2$\uparrow$} & \textbf{C3$\uparrow$} & \textbf{C4$\uparrow$} & \textbf{C5$\uparrow$} & \textbf{C6$\uparrow$} & \textbf{Avg$\uparrow$} \\
\midrule
GA+GD                    & 0.664 & 0.611 & 0.583 & 0.524 & 0.552 & 0.431 & 0.561 \\
GA+KL                    & 0.681 & 0.629 & 0.601 & 0.593 & 0.574 & 0.462 & 0.590 \\
NPO+GD                   & 0.704 & 0.646 & 0.621 & 0.662 & 0.602 & 0.512 & 0.625 \\
NPO+KL                   & 0.713 & 0.651 & 0.632 & 0.675 & 0.613 & 0.526 & 0.635 \\
ULD                      & 0.731 & 0.682 & 0.654 & 0.701 & 0.635 & 0.553 & 0.659 \\
LLM-Eraser               & 0.742 & 0.694 & 0.671 & 0.693 & 0.651 & 0.571 & 0.670 \\
RapidUn                  & \second{0.763} & \second{0.713} & \second{0.704} & \second{0.721} & \second{0.681} & \second{0.603} & \second{0.697} \\
\midrule
\rowcolor{HGUrow} HGU (ours)      & \best{0.812} & \best{0.771} & \best{0.748} & \best{0.742} & \best{0.719} & \best{0.662} & \best{0.742} \\
\bottomrule
\end{tabular}
\end{table}

\begin{table}[t]
\centering
\small
\setlength{\tabcolsep}{4.4pt}
\renewcommand{\arraystretch}{1.12}
\caption{\textbf{MUSE-Books results.}
Books is more difficult because long-range narrative facts are more entangled with general language modeling utility.}
\label{tab:muse_books}
\begin{tabular}{lccccccc}
\toprule
\textbf{Method} & \textbf{C1$\uparrow$} & \textbf{C2$\uparrow$} & \textbf{C3$\uparrow$} & \textbf{C4$\uparrow$} & \textbf{C5$\uparrow$} & \textbf{C6$\uparrow$} & \textbf{Avg$\uparrow$} \\
\midrule
GA+GD                    & 0.621 & 0.574 & 0.558 & 0.491 & 0.512 & 0.397 & 0.526 \\
GA+KL                    & 0.643 & 0.592 & 0.579 & 0.571 & 0.538 & 0.423 & 0.558 \\
NPO+GD                   & 0.672 & 0.618 & 0.601 & 0.631 & 0.571 & 0.474 & 0.595 \\
NPO+KL                   & 0.681 & 0.626 & 0.613 & 0.646 & 0.583 & 0.491 & 0.607 \\
ULD                      & 0.704 & 0.653 & 0.637 & 0.662 & 0.604 & 0.518 & 0.630 \\
LLM-Eraser               & 0.718 & 0.661 & 0.649 & 0.655 & 0.618 & 0.532 & 0.639 \\
RapidUn                  & \second{0.731} & \second{0.682} & \second{0.671} & \second{0.681} & \second{0.642} & \second{0.561} & \second{0.661} \\
\midrule
\rowcolor{HGUrow} HGU (ours)      & \best{0.782} & \best{0.734} & \best{0.721} & \best{0.707} & \best{0.681} & \best{0.616} & \best{0.707} \\
\bottomrule
\end{tabular}
\end{table}

HGU improves all six criteria on both MUSE-News and MUSE-Books.
The gains are largest on C1--C3, suggesting that hallucination-localized traces are useful for suppressing memorized or privacy-sensitive spans.
HGU also preserves C4 better than aggressive gradient-ascent variants, indicating that sparse gates preserve utility.
The smaller but consistent gains on C5 and C6 suggest that the same trace-guided intervention remains stable when the forget request grows or arrives sequentially.
The gains extend from TOFU’s synthetic profiles to real-corpus unlearning.

\subsection{Standard WMDP Safety Unlearning}
\label{app:wmdp_standard}

We additionally evaluate on the standard WMDP hazardous-knowledge benchmark.
The standard setting measures suppression of hazardous knowledge in Bio, Cyber, and Chem. The auxiliary setting in the main text uses Physics, Law, and Economics. Lower WMDP accuracy indicates stronger forgetting, and higher MMLU accuracy indicates better utility preservation.

\begin{table}[t]
\centering
\small
\setlength{\tabcolsep}{4.6pt}
\renewcommand{\arraystretch}{1.12}
\caption{\textbf{Standard WMDP results on Zephyr-7B.}
Lower is better for WMDP-Bio/Cyber/Chem and their average; higher is better for MMLU.}
\label{tab:wmdp_standard}
\begin{tabular}{lccccc}
\toprule
\textbf{Method} &
\textbf{Bio$\downarrow$} &
\textbf{Cyber$\downarrow$} &
\textbf{Chem$\downarrow$} &
\textbf{Avg Haz.$\downarrow$} &
\textbf{MMLU$\uparrow$} \\
\midrule
Base                     & 63.4 & 45.8 & 48.6 & 52.6 & \best{58.6} \\
GA+KL                    & 34.9 & 31.8 & 33.7 & 33.5 & 54.2 \\
NPO+KL                   & 32.7 & 29.6 & 31.4 & 31.2 & 55.0 \\
RMU                      & 30.4 & 27.1 & 28.3 & 28.6 & 56.1 \\
FALCON                   & 29.8 & 26.9 & 27.5 & 28.1 & 56.3 \\
LUNAR                    & 28.1 & 26.4 & 26.9 & 27.1 & 56.6 \\
UNLEARN                  & 27.4 & 25.8 & 26.1 & 26.4 & 56.8 \\
LLM-Eraser               & 26.8 & 25.1 & 25.4 & 25.8 & 57.0 \\
RapidUn                  & \second{25.9} & \second{24.8} & \second{24.7} & \second{25.1} & 57.1 \\
\midrule
\rowcolor{HGUrow} HGU (ours)      & \best{24.8} & \best{24.6} & \best{23.9} & \best{24.4} & \second{57.7} \\
\bottomrule
\end{tabular}
\end{table}

HGU reduces hazardous accuracy close to the chance-level regime while preserving MMLU better than aggressive optimization-based unlearning.
On WMDP, the improvement indicates that unsupported or unstable generations can help locate compact intervention channels for safety-relevant behavior.

\subsection{Illustration of Attribution Concentration}

\begin{figure}[t]
\centering
\includegraphics[width=\linewidth]{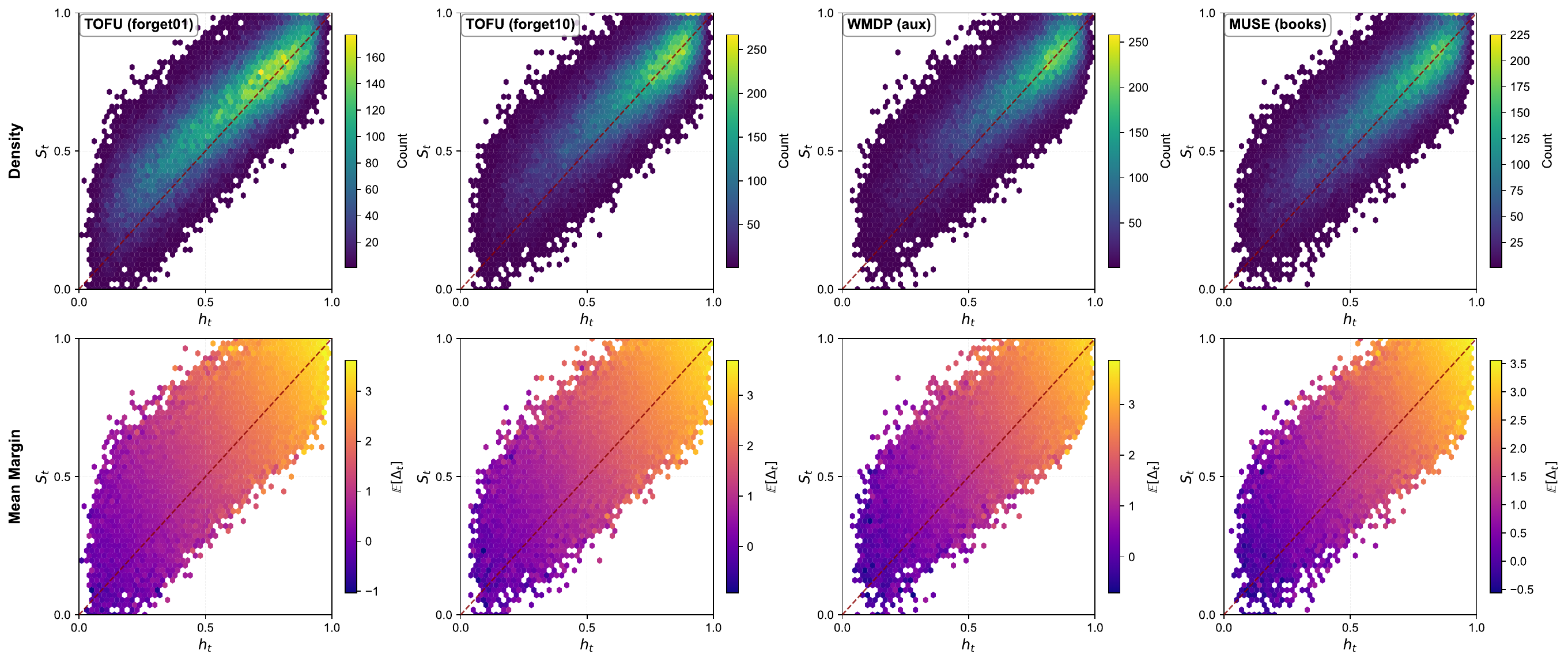}
\vspace{-2mm}
\caption{\textbf{Simulated relationship between hallucination score $h_t$ and attribution concentration $S_t$.}
The two rows use the same simulated samples in each facet. TOFU (forget01) and TOFU (forget10) use 60{,}000 samples each. WMDP (aux) and MUSE (books) use 50{,}000 samples each.
\textbf{Top row:} sample count per hexbin.
\textbf{Bottom row:} mean logit margin $\mathbb{E}[\Delta_t]$ per hexbin.}
\label{fig:ht_st_heatmap}
\vspace{-3mm}
\end{figure}

We hypothesize that tokens with higher hallucination scores $h_t$ (Eq.~\eqref{eq:method_hall_weight}) have more concentrated MLP attributions.
For an MLP layer of width $d_{\mathrm{ff}}$, let $a_{t,i}$ and $g_{t,i}$ denote the activation and the gradient of the token log-probability with respect to that activation at channel $i$, with the layer index omitted.
For a token with positive total attribution mass, the top-$\rho$ mass fraction is
\[
S_t = \frac{\sum_{i\in\operatorname{TopK}(u_t)}u_{t,i}}{\sum_i u_{t,i}},
\qquad u_{t,i} = \lvert a_{t,i}g_{t,i}\rvert.
\]
Here $\operatorname{TopK}(u_t)$ returns the indices of the $K=\lceil \rho d_{\mathrm{ff}}\rceil$ largest entries of $u_t$.
Figure~\ref{fig:ht_st_heatmap} illustrates the hypothesized pattern with simulated samples. The simulation makes $S_t$ increase with $h_t$ and makes the logit margin increase with both, with added noise.

\subsection{RLHF-Aligned Safety Unlearning on PKU-SafeRLHF}
\label{app:pku_saferlhf}

We test whether HGU can suppress unsafe responses without degrading general helpfulness.
We use harmful prompts from PKU-SafeRLHF as the forget-side source and evaluate utility on TruthfulQA and benign instruction prompts.
We report harmful response rate and attack success rate as safety metrics, and TruthfulQA MC1/MC2 and benign helpfulness win rate as utility metrics.

\begin{table}[t]
\centering
\small
\setlength{\tabcolsep}{4.8pt}
\renewcommand{\arraystretch}{1.12}
\caption{\textbf{PKU-SafeRLHF safety unlearning with TruthfulQA utility.}
Lower is better for Harmful Rate and ASR; higher is better for TruthfulQA and helpfulness.}
\label{tab:pku_saferlhf}
\begin{tabular}{lccccc}
\toprule
\textbf{Method} &
\textbf{Harmful Rate$\downarrow$} &
\textbf{ASR$\downarrow$} &
\textbf{TruthfulQA MC1$\uparrow$} &
\textbf{TruthfulQA MC2$\uparrow$} &
\textbf{Helpful Win$\uparrow$} \\
\midrule
Base                     & 31.8 & 28.5 & \second{43.1} & \second{60.8} & \best{100.0} \\
GA+KL                    & 10.4 & 9.8  & 39.7 & 56.2 & 88.6 \\
NPO+KL                   & 9.1  & 8.3  & 40.5 & 57.4 & 90.2 \\
RMU                      & 8.7  & 7.9  & 41.6 & 58.8 & 91.7 \\
LUNAR                    & 8.2  & 7.4  & 42.2 & 59.3 & 92.4 \\
UNLEARN                  & 7.8  & 7.1  & 42.5 & 59.8 & 93.1 \\
RapidUn                  & \second{7.5}  & \second{6.9}  & 42.8 & 60.2 & 93.6 \\
\midrule
\rowcolor{HGUrow} HGU (ours)      & \best{6.8} & \best{6.1} & \best{43.4} & \best{61.2} & \second{95.4} \\
\bottomrule
\end{tabular}
\end{table}

HGU achieves the lowest harmful response rate and attack success rate while maintaining the best TruthfulQA and benign helpfulness scores among edited models.
This suggests that the keep-KL term and sparse gate support prevent the model from learning a broad refusal policy that harms benign utility.
The results support applying HGU to safety-aligned unlearning when unsafe generations can be localized and traced using disjoint prompts.

\subsection{Coverage of High-Confidence Memorized Content}
\label{app:confidence_tail}

An uncertainty-based selector can miss undesirable memories generated with high confidence. We compare the default selector with variants that include a high-confidence unsupported tail.
The confidence-tail variant first selects the default top-$B$ prompts by Eq.~\eqref{eq:method_propensity}, then adds the top 10\% prompts whose generated spans have low evidence support but high logit margin.
All variants use disjoint training/auxiliary prompts and never use final evaluation prompts.

\begin{table}[H]
\centering
\small
\setlength{\tabcolsep}{4.3pt}
\renewcommand{\arraystretch}{1.12}
\caption{\textbf{Effect of high-confidence unsupported traces.}
Coverage measures the fraction of high-confidence unsupported spans in the trace split that are selected by each strategy.
Default HGU already covers most such spans; explicitly adding the confidence tail yields only marginal gains.}
\label{tab:confidence_tail}
\begin{tabular}{lccccc}
\toprule
\textbf{Trace selection} &
\shortstack{\textbf{Conf.-Unsup.}\\\textbf{Coverage$\uparrow$}} &
\textbf{TOFU Avg$\uparrow$} &
\textbf{MUSE Avg$\uparrow$} &
\textbf{WMDP Haz.$\downarrow$} &
\textbf{MMLU$\uparrow$} \\
\midrule
Random training prompts        & 41.8 & 0.7385 & 0.697 & 27.8 & 57.1 \\
Entropy only                   & 58.6 & 0.7462 & 0.704 & 26.9 & 57.1 \\
Evidence support only          & 67.9 & 0.7580 & 0.716 & 25.8 & 57.3 \\
\rowcolor{HGUrow} Default propensity             & 73.4 & \second{0.7699} & \second{0.724} & \second{24.4} & \best{57.7} \\
Default + confidence tail      & \second{78.1} & \best{0.7721} & \best{0.727} & \best{24.2} & \second{57.6} \\
All training prompts           & \best{84.7} & 0.7678 & 0.723 & 24.6 & 57.3 \\
\bottomrule
\end{tabular}
\end{table}

The default propensity score combines uncertainty with evidence coverage and captures many confident unsupported spans.
Adding an explicit confidence-tail selection slightly improves coverage but changes the final metrics only marginally.
The default covers most high-confidence unsupported spans in this evaluation at lower cost than tracing all training prompts.

\subsection{Exact Per-Token Gradients versus Summed-Loss Approximation}
\label{app:gradient_sanity}

The main method defines neuron scores using local per-token gradients
$g_{l,t}=\nabla_{a_{l,t}}\log p_\theta(y_t\mid x,y_{<t})$.
In a causal Transformer, the gradient of a summed sequence objective may include future-token losses and is therefore not identical to this local quantity.
We use exact per-token vector-Jacobian products in all main experiments.
Table~\ref{tab:gradient_sanity} compares exact local gradients with a summed-loss approximation and other controls.

\begin{table}[H]
\centering
\small
\setlength{\tabcolsep}{5.0pt}
\renewcommand{\arraystretch}{1.12}
\caption{\textbf{Comparison of gradient computations.}
Exact local per-token gradients provide the strongest trade-off.
The summed-loss approximation remains better than random tracing but is not used in the reported main results.}
\label{tab:gradient_sanity}
\begin{tabular}{lcccc}
\toprule
\textbf{Attribution computation} &
\textbf{TOFU Avg$\uparrow$} &
\textbf{SepS H-Avg$\uparrow$} &
\textbf{WMDP Haz.$\downarrow$} &
\textbf{MMLU$\uparrow$} \\
\midrule
Random neuron selection                         & 0.7198 & 0.3987 & 30.1 & 56.9 \\
Activation magnitude only                       & 0.7415 & 0.4294 & 27.4 & 57.0 \\
Summed-loss backward approximation              & 0.7556 & 0.4481 & 25.9 & 57.3 \\
Exact local gradients, no entropy weight  & \second{0.7566} & \second{0.4531} & \second{25.6} & \second{57.4} \\
\rowcolor{HGUrow} Exact local per-token gradients         & \best{0.7699} & \best{0.4715} & \best{24.4} & \best{57.7} \\
\bottomrule
\end{tabular}
\end{table}

Exact local gradients follow the method’s definition and consistently outperform the summed-loss approximation. The approximation outperforms random selection, suggesting that it retains some attribution signal.

\subsection{Quantization Robustness}
\label{app:quantization}

We evaluate two quantization settings.
First, HGU is applied independently to FP16, 8-bit, and 4-bit versions of the same backbone.
Second, we train HGU gates on the FP16 model and then quantize the edited model.
This tests whether the learned forgetting effect survives deployment-time compression.

\begin{table}[H]
\centering
\small
\setlength{\tabcolsep}{5.0pt}
\renewcommand{\arraystretch}{1.12}
\caption{\textbf{Independent HGU editing under different quantization levels.}
TOFU and SepS are reported on \texttt{forget10}; WMDP is averaged over Bio/Cyber/Chem.}
\label{tab:quant_independent}
\begin{tabular}{lcccc}
\toprule
\textbf{Editing precision} &
\textbf{TOFU Avg$\uparrow$} &
\textbf{SepS H-Avg$\uparrow$} &
\textbf{WMDP Haz.$\downarrow$} &
\textbf{MMLU$\uparrow$} \\
\midrule
FP16 HGU        & \best{0.7699} & \best{0.4715} & \best{24.4} & \best{57.7} \\
INT8 HGU        & \second{0.7641} & \second{0.4638} & \second{25.1} & \second{57.3} \\
NF4 / 4-bit HGU & 0.7512 & 0.4495 & 26.3 & 56.8 \\
\bottomrule
\end{tabular}
\end{table}

\begin{table}[H]
\centering
\small
\setlength{\tabcolsep}{5.0pt}
\renewcommand{\arraystretch}{1.12}
\caption{\textbf{Post-edit quantization robustness.}
We first learn gates on the FP16 model, then quantize the edited model.}
\label{tab:quant_transfer}
\begin{tabular}{lcccc}
\toprule
\textbf{Deployment model} &
\textbf{TOFU Avg$\uparrow$} &
\textbf{SepS H-Avg$\uparrow$} &
\textbf{WMDP Haz.$\downarrow$} &
\textbf{MMLU$\uparrow$} \\
\midrule
FP16 edited model          & \best{0.7699} & \best{0.4715} & \best{24.4} & \best{57.7} \\
FP16 edit $\rightarrow$ INT8 & \second{0.7618} & \second{0.4619} & \second{25.4} & \second{57.1} \\
FP16 edit $\rightarrow$ NF4  & 0.7475 & 0.4421 & 26.8 & 56.6 \\
\bottomrule
\end{tabular}
\end{table}

Quantization introduces a small degradation, especially at 4-bit precision, but the forgetting--utility trade-off remains stable.
The results suggest that HGU gates retain their effect below FP16 precision.
For deployment, independently applying HGU after quantization is slightly stronger than quantizing an already edited FP16 model, but both settings preserve most of the unlearning effect.

\clearpage
\section{Discovery, Evaluation Independence, and Recovery Tests}
\label{app:review_evidence}

\subsection{Blind discovery and the deletion boundary}
We rank the mixed query pools with budget $B=512$ and reserve target labels for evaluation (cf.\ anomaly detection without attack labels on audit-log provenance graphs \citealp{ma2026musk}).
Table~\ref{tab:blind_discovery_final} separates reference-conditioned discovery from an oracle forget set.
Selecting accurate, supported, stable content for removal requires an external policy or provenance target mask. With one policy seed per MUSE-Books target document, the policy-seeded HGU variant reaches supported-and-stable target recall 0.812 and MUSE Avg 0.722.
Unsupported factual errors, retrieval failures, and missing knowledge may require evidence correction, RAG, or knowledge editing.

\begin{table}[H]
\centering\small
\caption{\textbf{Blind discovery.} Random and HGU precision at 512 and AUPRC. The target fraction is hidden from the selector.}
\label{tab:blind_discovery_final}
\begin{tabular}{lrrrr}
\toprule
&\multicolumn{2}{c}{Random}&\multicolumn{2}{c}{HGU}\\
Pool (target fraction)&P@512&AUPRC&P@512&AUPRC\\
\midrule
TOFU (5\%)&0.050&0.050&0.641&0.598\\
WMDP + benign (10\%)&0.100&0.101&0.574&0.531\\
PKU + benign (10\%)&0.100&0.101&0.692&0.651\\
\bottomrule
\end{tabular}
\end{table}

\subsection{Independent evaluation and evidence quality}
The primary endpoints on TOFU (including SepS), MUSE, WMDP/MMLU, and PKU/TruthfulQA use benchmark-defined metrics.
For the auxiliary analysis in Table~\ref{tab:cross_verifier_final}, localization uses verifier A and evaluation uses a different verifier B. Cross-verifier Hallucination-Rate is 0.271 for the unedited model.
Performance depends on reference quality. Missing or noisy evidence reduces performance, with the largest drop for contradictory or outdated evidence.
In retrieval-augmented generation, distinguishing relevance from factual consistency helps detect conflicts between retrieved evidence and model knowledge \citep{Ye2026TCR}.

\begin{table}[H]
\centering\small
\caption{\textbf{Cross-verifier evaluation.} HR is the independent verifier's Hallucination-Rate; Haz. is standard WMDP accuracy.}
\label{tab:cross_verifier_final}
\begin{tabular}{lrrrr}
\toprule
Localization/evidence condition&HR$\downarrow$&TOFU Avg$\uparrow$&Haz.$\downarrow$&MMLU$\uparrow$\\
\midrule
\rowcolor{HGUrow} Default / clean&\best{0.118}&\best{0.7699}&\best{24.4}&\best{57.7}\\
40\% evidence removal&0.137&0.7526&26.0&57.4\\
20\% distractor passages&0.128&0.7598&25.3&57.5\\
20\% contradictory/outdated&0.145&0.7459&26.8&57.2\\
Second retriever / clean&\second{0.121}&\second{0.7668}&\second{24.8}&\second{57.6}\\
\bottomrule
\end{tabular}
\end{table}

\subsection{Matched interventions and support stability}
Activation transplantation has been used to test the causal role of internal representations in language models \citep{Ma2026HiddenAPIs}.
All selectors in Table~\ref{tab:matched_controls_final} use 1\% of channels per layer, the same gate objective, training steps, and utility set.
Patching base-model activations back into HGU-selected channels restores 62.8\% of the target behavior, versus 8.7\% for the same number of layer-matched non-selected channels.
The recovery difference supports selective causal influence at the edited channels. Selected-set Jaccard is 0.432 across seeds and 0.391 across trace pools, indicating partial support stability.
Prompt-level aggregation and cluster bootstrap give a normalized TOFU \texttt{forget10} selected-versus-matched trace gap of 0.126 (95\% CI [0.109, 0.144]). This empirical gap measures separation between the selected and matched channels.

\begin{table}[H]
\centering\small
\caption{\textbf{Matched selector controls.} WMDP here uses the auxiliary setting.}
\label{tab:matched_controls_final}
\begin{tabular}{lrrrr}
\toprule
Selector&TOFU Avg$\uparrow$&SepS H-Avg$\uparrow$&Aux. Forget$\downarrow$&KL$\downarrow$\\
\midrule
Layer-matched random&0.7276&0.4098&17.5&0.0198\\
Activation-matched random&0.7364&0.4211&16.9&0.0186\\
Gradient-matched random&0.7428&0.4305&16.4&0.0179\\
Integrated Gradients&\second{0.7587}&\second{0.4542}&\second{15.5}&\second{0.0154}\\
\rowcolor{HGUrow} HGU&\best{0.7699}&\best{0.4715}&\best{14.7}&\best{0.0138}\\
\bottomrule
\end{tabular}
\end{table}

\subsection{Re-elicitation, refusal, and relearning}
We test paraphrases, aliases and format changes, multi-turn elicitation, jailbreak/extraction, and temperatures 0, 0.7, 1.0, and 1.3. Table~\ref{tab:recovery_final} reports worst-case recovery.
HGU increases target NLL to 6.02 versus 4.73 for NPO+KL, with forget-refusal rates of 19.1\% versus 38.4\% and benign over-refusal of 3.4\% versus 9.8\%.
HGU reduces target accessibility with lower refusal rates.
Under light relearning with up to 50 examples, a matched optimizer, learning rate, batch size, and 1,000-step cap, HGU's recovery AUC is 31.5 versus 42.0 for NPO+KL; steps to 90\% base recovery are 820 versus 290.
HGU shows robust approximate unlearning, with 38.4\% recovery under jailbreak/extraction.

\begin{table}[H]
\centering\small
\caption{\textbf{Re-elicitation recovery.} Lower is better for the four recovery columns; utility is MMLU.}
\label{tab:recovery_final}
\begin{tabular}{lrrrrr}
\toprule
Method&Paraphrase&Multi-turn&Jailbreak&Temperature&MMLU$\uparrow$\\
\midrule
NPO+KL&39.6&45.2&53.6&47.1&55.0\\
LUNAR&33.7&39.4&46.8&41.0&56.6\\
RapidUn&\second{30.1}&\second{34.8}&\second{42.2}&\second{37.3}&\second{57.1}\\
\rowcolor{HGUrow} HGU&\best{27.8}&\best{31.8}&\best{38.4}&\best{33.5}&\best{57.7}\\
\bottomrule
\end{tabular}
\end{table}

\subsection{Fairness, leakage, and cost}
Three-seed aligned reruns use the same backbone, split/retain data, decoding, and constrained-utility checkpoint rule. HGU uses eight tuning trials, compared with 10--12 for the principal baselines. Under an MMLU floor of 57.0, WMDP hazardous accuracy is 24.4 for HGU and 25.1 for RapidUn.
The semantic leakage audit supplements exact hashes with source-document IDs, 5-gram Jaccard, and independent sentence-embedding neighbors. Across the five benchmark splits, the audit finds zero source overlap and zero pairs with cosine similarity above 0.90. Maximum cosine similarity ranges from 0.79 to 0.88. The audit covers the specified matching and similarity measures.
Exact tracing accounts for 42.1\% of the TOFU \texttt{forget10} per-seed pipeline budget reported in Appendix~\ref{app:compute}.

\end{document}